\pdfoutput=1
\documentclass[11pt]{article}

\usepackage{acl}

\usepackage{times}
\usepackage{latexsym}

\usepackage[T1]{fontenc}
\usepackage[utf8]{inputenc}

\usepackage{microtype}

\usepackage{inconsolata}

\usepackage{graphicx}
\usepackage{caption}
\usepackage{times}
\usepackage{latexsym}
\usepackage{booktabs}
\usepackage{amssymb}
\usepackage{amsmath}
\usepackage{bbm}
\usepackage{mathtools}
\usepackage{url}
\usepackage{float}
\usepackage{multirow}
\usepackage{subcaption}
\usepackage{xcolor}
\usepackage{xspace}
\usepackage{ulem} 
\usepackage{soul}
\usepackage{svg}
\usepackage{listings}
\usepackage{makecell}
\newcommand{\model}{\textsc{PROF}\xspace}
\newcommand{\best}[1]{\colorbox[HTML]{d2e7d6}{\textbf{#1}}}
\newcommand{\secondbest}[1]{\colorbox[HTML]{c0e7f6}{#1}}
\title{Closing the Loop: Learning to Generate Writing Feedback \\via Language Model Simulated Student Revisions}

\author{Inderjeet Nair$^{\alpha}$, Jiaye Tan$^{\alpha}$, Xiaotian Su$^{\beta}$, Anne Gere$^{\alpha}$, Xu Wang$^{\alpha}$, Lu Wang$^{\alpha}$\\
  $^{\alpha}$University of Michigan, $^{\beta}$ETH Zürich \\
  \texttt{inair@umich.edu} 
  }

\begin{document}
\maketitle

\begin{abstract}
% While the act of providing feedback is widely recognized as a crucial practice for refining student writing and deepening their understanding, the challenge lies in crafting feedback that can be effectively implemented to maximize improvement in the initial content. 
Providing feedback is widely recognized as crucial for refining students' writing skills. %and deepening their understanding of the writing subjects. %However, the challenge lies in crafting timely, high-quality feedback that can be effectively implemented by the students to improve upon the initial content. 
Recent advances in language models (LMs) have made it possible to automatically generate feedback that is actionable and well-aligned with human-specified attributes. 
However, it remains unclear whether the feedback generated by these models is truly effective in enhancing the quality of student revisions. 
Moreover, prompting LMs with a precise set of instructions to generate feedback is nontrivial due to the lack of consensus regarding the specific attributes that can lead to improved revising performance. 
To address these challenges, we propose \model that \ul{PRO}duces \ul{F}eedback via learning from LM simulated student revisions. \model aims to iteratively optimize the feedback generator by directly maximizing the effectiveness of students' overall revising performance as simulated by LMs. 
Focusing on an economic essay assignment, we empirically test the efficacy of \model and observe that our approach not only surpasses a variety of baseline methods in effectiveness of improving students' writing but also demonstrates enhanced pedagogical values, even though it was not explicitly trained for this aspect.

% Moreover, prompting LMs with a precise set of instructions to generate feedback is a complex task due to the lack of consensus regarding the specific attributes it should encompass to maximize student implementation performance.
% Additionally, the conditions that influence feedback effectiveness are highly complex \lu{condition on student?} and difficult to articulate \lu{i'd not say it's hard to articulate, but rather nuanced, complex, becuz experts should be able to tell}, rendering current alignment techniques less directly applicable.\lu{not sure if we want to talk about alignment technique here, a bit too early, it should be part of your proposal} 
% To address these challenges, we propose \model (\ul{PRO}duces \ul{F}eedback via learning from simulated revisions) that models student revision process and uses it to iteratively fine-tune the feedback generation system. Our goal is to optimize this system so that it maximizes the effectiveness of student revisions and enhances their overall performance. We empirically test the efficacy of \model on student-written essays for an assignment and observe that our approach not only surpasses a variety of baseline methods in effectiveness but also demonstrates enhanced pedagogical value, even though it was not explicitly trained for this aspect.

\end{abstract}

\section{Introduction}
Writing high-quality essays often requires subject-specific and customized feedback from peers and experts, followed by multiple rounds of revisions~\cite{fitzgerald1987teaching,hayes1987cognitive,macarthur1991knowledge,afrin-litman-2023-predicting}. 
As students incorporate feedback into their writing, they not only improve the current piece but also advance the general writing skills, learn to critically self-assess their work~\cite{macarthur2007best}, and gain a deeper understanding of the subject matter~\cite{bangert2004effects}.

% Writing good quality essays requires multiple iterations of high-quality feedback from peers, experts, or automated feedback generation systems, followed by refinements based on the assessment~\cite{fitzgerald1987teaching,hayes1987cognitive,macarthur1991knowledge,afrin-litman-2023-predicting}. 
% As students incorporate feedback into their initial writing, they not only improve the current piece but also advance their general writing skills, learn to critically self-assess their work~\cite{macarthur2007best}, and gain a deeper understanding of the subject matter~\cite{bangert2004effects}.

Recent advances in language models (LMs)~\cite{hoffmann2022training,chowdhery2023palm,touvron2023llama,jiang2024mixtral} make it possible to develop automatic feedback generation systems to provide concrete and actionable comments in a timely manner~\cite{chamoun2024automated,d2024marg}, compared to the time-consuming process performed by humans. 
However, careful prompt engineering is necessary to incorporate precise instructions, ensuring that the generated feedback effectively guides students in improving the quality of their writing. 
More importantly, providing such detailed instructions is not a trivial task since there is still no general consensus about what attributes the feedback must entail to effectively contribute to students' learning outcomes~\cite{nelson2009nature}. 
For example, \citet{bitchener2005effect} show that including explanations in feedback can only improve the writing quality of specific revisions, and sometimes (e.g., if too lengthy) can negatively affect tenth-graders' overall writing performance~\cite{tseng2007line}. 
% personalized
% Furthermore, the attributes of effective feedback vary depending on the students' writing ability levels and their understanding of the subject matter~\cite{underwood2006improving}.

% Owing to the recent advances in large language models (LLMs)~\cite{hoffmann2022training,chowdhery2023palm,touvron2023llama,jiang2024mixtral}, it has been possible to build helpful feedback generation systems capable of generating specific and actionable comments via prompting~\cite{chamoun2024automated,d2024marg}. However, the effectiveness of such feedback in maximizing student implementation performance remains uncertain and requires further investigation. 
% One critical challenge associated with the above prompting-based methodologies is their reliance on precise instructions from humans to capture the essential elements necessary for feedback to effectively yield quality improvement during refinement. 
% It is difficult to articulate these instructions as there is no general consensus about what attributes the feedback must entail so that it can be best implemented by the students~\cite{nelson2009nature}. For example, while \citet{bitchener2005effect} found that including explanations in feedback led to improved performance in the revised version, tenth-graders' performance in writing drafts was negatively affected by the lengthy explanations~\cite{tseng2007line}. 
% Furthermore, the attributes for effective feedback vary depending on the students' ability level and subject matter understanding~\cite{underwood2006improving}.

To this end, our goal is to build an automatic feedback generation system that can be \textit{directly optimized to maximize students' writing revision performance}, to avoid the complexity of explicitly specifying the criterion for effective feedback. 
However, involving actual students at every stage of the system-building process is impractical due to the time required and the potential negative impact on participants from an immature system~\cite{latifi2021does}. 
To address this challenge, we first develop an \textbf{LM-based student simulator} that emulates the process of applying feedback to revise initial content, inspired by the recent efforts to simulate human processes~\cite{10.1145/3586183.3606763,shanahan2023role,xu2023leveraging,lu2024generative}. In our empirical evaluations of LM simulators, we discovered that by varying the temperature used in autoregressive decoding, we can effectively simulate a diverse array of behaviors to support a comprehensive testing of the feedback generator. 
% This outcome is advantageous as it allows us to comprehensively test the performance of different feedback generation systems across a wide range of personas.

We then propose a feedback generation model, \textbf{\model}\footnote{Our code and data are available at \url{https://github.com/launchnlp/PROF}.}, that \ul{PRO}duces \ul{F}eedback via learning from LM-simulated student revisions. 
Concretely, we use the LM student simulator to iteratively generate preference relations involving desirable and undesirable feedback. We then apply the Direct Preference Optimization (DPO) objective~\cite{NEURIPS2023_a85b405e,xu2023some,yuan2024self,pang2024iterative} along with the preference relations to update the feedback generator. Importantly, the iterative process aims to enhance the effectiveness of the generated feedback, resulting in better implementation performance according to the student simulator.

To evaluate the feedback generated by \model, we conduct a study on an essay assignment from an introductory economics course offered at a university in United States. In our experimental analyses, we compared the performance of our model with that of few-shot prompted \texttt{gpt-3.5/gpt-4} models along with other nontrivial comparisons. 
Our approach not only achieves a similar level of pedagogical alignment, but also outperforms these enterprise LMs in terms of implementation performance of essay revising. Notably, our model is significantly smaller in size ($8$ billion parameters), making it more efficient and can be easily adapted to other writing assignments. 
Furthermore, in our empirical experiments involving the student simulators, we observed that the feedback generated by our model aligns well with the actual human revisions, demonstrating its faithfulness in implementation.

To summarize, our work makes the following major contributions: 
% \begin{itemize}
%     \item We propose \model, a method that trains feedback generation models by eliminating the need for manually defining the desired feedback attributes or relying on any large-scale annotated dataset of high-quality feedback. 
%     \item We propose a method for automatically evaluating the effectiveness of the generated feedback in terms of student implementation performance using LM-based student simulators.
%     \item By tuning the temperature involved in auto-regressive decoding, we can generate a wide-range of behaviour from the student simulator allowing us to subsequently develop a feedback generation system that caters to diverse writing traits. 
%     This versatility can be harnessed to customize feedback generators and optimize implementation performance for distinct behaviors. 
%     %as well as create testing environments to evaluate the performance of feedback generators across a range of diverse personas.
    
% \end{itemize}

\begin{itemize}
    \item We propose \model, a method that trains feedback generation models by eliminating the need for manually defining the desired feedback attributes or relying on any large-scale annotated dataset of high-quality feedback. 

    \item We propose a method for automatically evaluating the effectiveness of the generated feedback in terms of student implementation performance using LM-based student simulators. 

    \item By tuning the temperature involved in auto-regressive decoding, we can generate a wide-range of behaviour from the student simulator allowing us to subsequently develop a feedback generation system that caters to diverse writing traits. This versatility can be harnessed to customize feedback generators and optimize implementation performance for distinct behaviors.

\end{itemize}

% This allows us to enjoy the following advantages: \textbf{(1)} Our approach eliminates the need for manually defining the desired attributes of feedback or relying on a large-scale dataset of high-quality feedback annotations. \textbf{(2)} Although our technique optimizes the feedback generation LLM to maximize student implementation performance, it also results in the generation of feedback that are more pedagogically aligned. \textbf{(3)} By tuning the temperature involved in auto-regressive decoding, we can generate a wide-range of behaviour from the student simulator allowing us to subsequently develop a feedback generation system that caters to diverse personas.

\begin{figure*}
    \centering
    \includegraphics[scale=0.45]{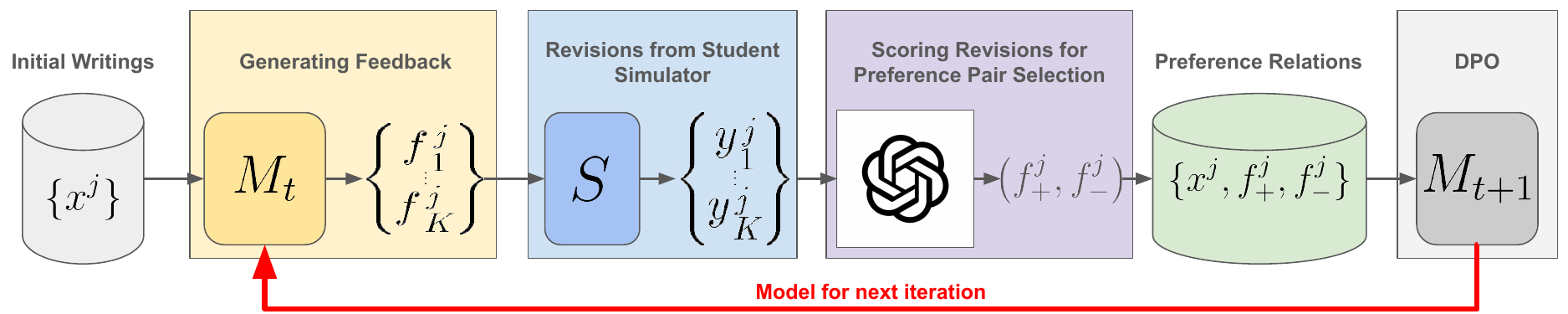}
    \caption{
    % \lu{this fig should appear early, e.g. on p2 or p3}
    \textbf{\model Pipeline}: The depicted figure illustrates the iterative optimization algorithm used in our approach. At each iteration $t$, the feedback generator $M_{t}$ generates multiple feedback samples, which are then evaluated for their effectiveness using the student simulator and then \texttt{gpt-4} as a judge. These evaluations are used to establish preference relations over feedback using the quality of the corresponding revised essays
    % \lu{over feedback? as judged by student revision quality?}
    , which are subsequently used to update the parameters of $M_{t}$ via DPO~\cite{NEURIPS2023_a85b405e}, resulting in the updated version $M_{t + 1}$.}
    \label{fig:prof_pipeline}
\end{figure*} 

\section{Data Description}
\label{sec:dataset_description}
We collected data from the essay assignments submitted by students enrolled in the Economics 101 course at the University of Michigan, Ann Arbor, United States.  
This assignment explores a scenario in which ``\textit{an increase in the minimum wage in San Francisco could lead to burgeoning adoption of automation}''. To discourage this outcome, two policies are proposed: a) a tax on automation and b) a ban on automation. 
The students are instructed to craft a persuasive letter explaining the economic consequences of a minimum wage increase. Furthermore, they are tasked with presenting arguments against one of the aforementioned policies by using the tools and principles taught in the course.
% The students are instructed to craft a persuasive letter of $400$-$500$ words addressed to the San Francisco Board of Supervisors. The letter should explain the economic consequences of a minimum wage increase in the labor market, as well as its extended impact on the automation and goods markets. Furthermore, students are required to choose one of the policies and present counterarguments against it, utilizing the economic tools and principles they have acquired throughout the course. 
Refer Appendix \ref{sec:assignment_prompt} to view the assignment prompt.
% \lu{this para can be shortened if needed}

Thereafter, each student essay is reviewed by $3$ peers to obtain a set of $3$ feedback which identifies areas for improvement. 
% where the essay requires the most attention. 
This assignment uses scripted peer feedback~\cite{latifi2021does,noroozi2016relations} wherein the peer reviewers are expected to provide feedback along a series of prompts. Refer Appendix \ref{sec:peer_review_prompt} to view the questions in the scripted peer feedback. Finally, the author revises the original essay based on the received peer reviews.

In total, we collected $363$ datapoints, each comprising the \textbf{initial writing}, \textbf{three peer feedback}, and the \textbf{revised essay}. Among these, we utilized $291$ essay-revision pairs along with the $873\, (291 \times 3)$ feedback for our train split, while the remaining $72$ datapoints were used for testing. As demonstrated later, our approach yields enhanced performance despite the limited number of datapoints and the absence of expert annotated feedback. 
This assignment includes a detailed rubric for evaluating and grading student-written essays. Our approach utilizes this rubric to create an essay evaluation prompt for assessing revisions produced by LM student simulators. This prompt can be found in Appendix \ref{sec:grade_essay_prompt}.

\section{\model: Learning to Generate Feedback with Simulated Student Revisions}
\label{sec:model}

% \lu{comment out content, feel free to add back}
To directly optimize the feedback generation on student revising performance, we have two LMs that function as \textbf{feedback generator} and \textbf{student simulator} respectively, as illustrated in Figure~\ref{fig:prof_pipeline}. 
In \S \ref{sec:student_simulation}, we first describe the training of student simulators using the data from \S \ref{sec:dataset_description}, to emulate how students integrate feedback into revisions. 
In \S \ref{sec:initializing_feedback_generator} and \S \ref{sec:feedback_iterative_optimization}, we present how the feedback generator is initialized and iteratively optimized by the proposed \model method based on desirable and undesirable feedback, as measured by simulated revisions' quality. 

% In \S \ref{sec:student_simulation}, we describe the training process of our student simulator using the dataset described in \S \ref{sec:dataset_description}. This simulator is specifically designed to emulate how students integrate feedback into their initial content to generate the revised content. 
%
% Thereafter, in \S \ref{sec:initializing_feedback_generator} and \S \ref{sec:feedback_iterative_optimization},
% we describe how the feedback generator is initialized and the details of iterative preference optimization respectively. 
% On a high level, we generate multiple feedback instances from the feedback generator in each iteration. These instances are then evaluated based on the quality of the corresponding revisions produced by the student simulator. This evaluation process creates preference relations that highlight desirable and undesirable feedback. To align the feedback generator, we employ a preference optimization technique such as DPO~\cite{NEURIPS2023_a85b405e}. 
% Figure \ref{fig:prof_pipeline} pictorially represents the optimization strategy described in this section. In this way, our algorithm offers a strategy to explicitly optimize feedback generation in terms of student revision performance, without the need for a specific criterion that defines effective feedback.

\subsection{Student Simulation}
\label{sec:student_simulation}
% In this section, we explain how the data described in \S \ref{sec:dataset_description} can be used to simulate the behavior of the students for the task of implementing feedback.

We represent the dataset of the assignment submissions as $\mathcal{D}$, whose $j^{\text{th}}$ element can be represented as a tuple $(x^j, \{f^j_i\}_{i=1}^3, y^j)$. Here, $\{f_i^j\}_{i=1}^3$ represents the set of feedback applied by the student to revise the initial writing $x^j$ into $y^j$. 

To simulate the behavior of implementing one feedback in place of $3$ feedback simultaneously, we instruct \texttt{gpt-3.5}\footnote{\url{https://platform.openai.com/docs/models/gpt-3-5-turbo}} to combine the $3$ feedback into a holistic feedback $f^j_c$. Please refer Appendix \ref{sec:combining_feedback_prompt} to view the exact prompt.

% \lu{this para too long, remove. just say we concate the three feedback into one then fine-tune two llms ... leave details to appendix}
% If we directly optimize the model to generate $y^j$ given $x^j$ and $\{f^j_i\}_{i=1}^3$, it would simulate the behavior of implementing \textit{multiple} feedback simultaneously. However, in our iterative algorithm, we aim to assign a score to each feedback generated by the feedback generator individually, in order to construct preference relations. Therefore, we combine the $\{f^j_i\}_{i=1}^3$ into a holistic feedback $f^j_c$ that captures all the comments from the individual feedback using \texttt{gpt-3.5}\footnote{\url{https://platform.openai.com/docs/models/gpt-3-5-turbo}} via few-shot prompting. The prompt used for combining feedback can be found in the Appendix \ref{sec:combining_feedback_prompt}.

Thereafter, we fine-tune two LMs, $\texttt{llama3-8b}$\footnote{\url{https://llama.meta.com/llama3}} and $\texttt{gpt-3.5}$, of different scales to implement the feedback, i.e., generating $y^j$ given $x^j$ and $f^j_c$. 
% Note that the combined feedback $f^j_c$ is solely used as a component of the input during the training process of the student simulator. 
% As we will explore in more detail in \S \ref{sec:initializing_feedback_generator}, the actual individual peer reviews are employed as the fine-tuning data when initializing the feedback generator.
We represent the trained simulator as $S$ and use it to sample revisions during feedback generator training for initial writing $x$ and feedback $f$, i.e., $y \sim S(\cdot | x, f)$. 

\subsection{Feedback Generator: Initialization}
\label{sec:initializing_feedback_generator}
We initialize our feedback generator using \texttt{llama3-8b} and train it specifically for the task of generating peer feedback.
To create paired data for fine-tuning, we use $\mathcal{D}$ where the $j^{\text{th}}$ assignment submission consists of three paired data points $\{x^j, f^j_i\}_{i=1}^3$ using the individual peer feedback.
We represent the feedback generator by $M$ from which the feedback $f$ can be sampled for an essay $x$ as $f \sim M(\cdot | x)$. After this initialization process, we continue training using the \model method to optimize student revision performance

% \subsection{Feedback Generator: Iterative Optimization}
\subsection{Feedback Generator: Optimization}
\label{sec:feedback_iterative_optimization}
Our approach assumes access to two functions for the iterative optimization of the feedback generator: (a) student simulator as described in \S \ref{sec:student_simulation} and (b) automatic essay scoring system, where we use \texttt{gpt-4} via few shot prompting owing to their strong capabilities in critically assessing the quality of natural language outputs~\cite{zheng2023lmsys,li2023generative,zheng2024judging}. 
The rubrics employed to assess the quality of the essay are identical to the course rubrics. Refer to Appendix \ref{sec:grade_essay_prompt} to view the complete prompt.
% While \S \ref{sec:student_simulation} describes how student simulator is set up, 
% We use \texttt{gpt-4} via few shot prompting to automatically score essays owing to their strong capabilities in critically assessing the quality of natural language outputs and their alignment with respect to certain attributes~\cite{zheng2023lmsys,li2023generative,zheng2024judging}. 
% The rubrics employed to assess the quality of the essay are identical to the course rubrics used for grading the essays, as discussed in Section \ref{sec:dataset_description}. Please refer to Appendix \ref{sec:grade_essay_prompt} to view the complete prompt.

Let the feedback generator at the start of $t^{th}$ iteration be represented by $M_{t}$. Our objective would be to use $M_{t}$ in creating desirable feedback $f^j_{+}$ and undesirable feedback $f^j_{-}$ for each initial essay $x^j$ in $\mathcal{D}$. After constructing such kind of preference relationship for each datapoint, we use Direct Preference Optimization (DPO) loss~\cite{NEURIPS2023_a85b405e} to train a new model $M_{t + 1}$ for the next iteration. The following objective describes the relation between $M_{t + 1}$ and $M_{t}$ using the DPO loss:

\begin{equation}
\label{eqn:dpo_objective}
\scriptsize
M_{t+1} = \arg \min_{M_\theta}\left[\sum_{j=1}^{\mathcal{|D|}} \mathcal{L}_t(f^j_+, f^j_-, x)\right]
\end{equation}
\begin{equation}
\label{eqn:dpo_loss}
\scriptsize
\mathcal{L}_t(f^j_+, f^j_-, x^j) =  -\log \sigma\left(\beta \frac{M_\theta(f^j_+|x^j)}{M_{t}(f^j_+|x^j)} - \beta \frac{M_\theta(f^j_-|x^j)}{M_{t}(f^j_-|x^j)}\right)
\end{equation}
% \lu{eq2 x doesn't have index j, if you don't have Mt on the left, can remove M theta too. two eqs not the same size}
The feedback generator $M_\theta$ is initialized with the parameters of $M_{t}$ and after optimizing Eq.~\ref{eqn:dpo_objective}, it becomes $M_{t + 1}$. The loss in Eq.~\ref{eqn:dpo_loss} enables the model to effectively discern high quality feedback from the low quality ones by amplifying the difference in likelihood between the two, relative to the likelihood estimates of $M_{t}$. 
%desirable and undesirable feedback, relative to the likelihood estimates of $M_{t}$.
% \lu{what's this? introduced yet?} and $M_{t - 1}$ are used for probability estimation. \lu{give an intuitive explanation on what does this objective do, e.g. making the model discern low quality feedback by comparing it with high quality ones, etc}

To generate the preference pair $(f^j_{+}, f^j_{-})$ for each initial writing $x$ at iteration $t$, we apply the following steps in sequence:
\begin{enumerate}
    \item Sampling $K$ different feedback from $M_{t}$ for each datapoint $x^j$ as: $\{f^j_k\}_{k=1}^K \sim M(\cdot | x^j)$ .
    % using temperature sampling\lu{what does this mean? which temp do you use? it also varies?}.
    \item The trained student simulator $S$ is used to generate the revised version for each of the generated feedback in $\{f^j_k\}_{k=1}^K$.
    \item Finally, the reward for each feedback is estimated by employing \texttt{gpt-4} as an evaluator that assesses the quality of the revised essay for each feedback. The feedback associated with the best revision and the worst revision is chosen as the preference pair. 
    % \lu{can stop here} highlighting desirable and undesirable feedback.
\end{enumerate}

It is important to note that our algorithm relies on minimal supervision from \texttt{gpt-4}, which is solely used for evaluating the quality of the generated essays. 
% Furthermore, to reduce compute costs, an alternative approach would be training a more targeted and compact model for automatic essay scoring instead of relying on \texttt{gpt-4}.\lu{2nd sent can remove, whole para can remove too}

% \begin{figure*}[h]
%      \centering
     
%         % \caption{
%         % % \lu{make all the figures color-blind friendly, use different shapes or at least avoid using green and red; text in the figure should be bigger. these comments apply to other figures too}
%         % \lu{can only keep 1 fig, the other goes appendix}
%         % Number of lexical modifications at different temperatures.}
%         \label{fig:num_modifications}
% \end{figure*}

\begin{figure}[t]
    \centering
    \includegraphics[scale=0.3]{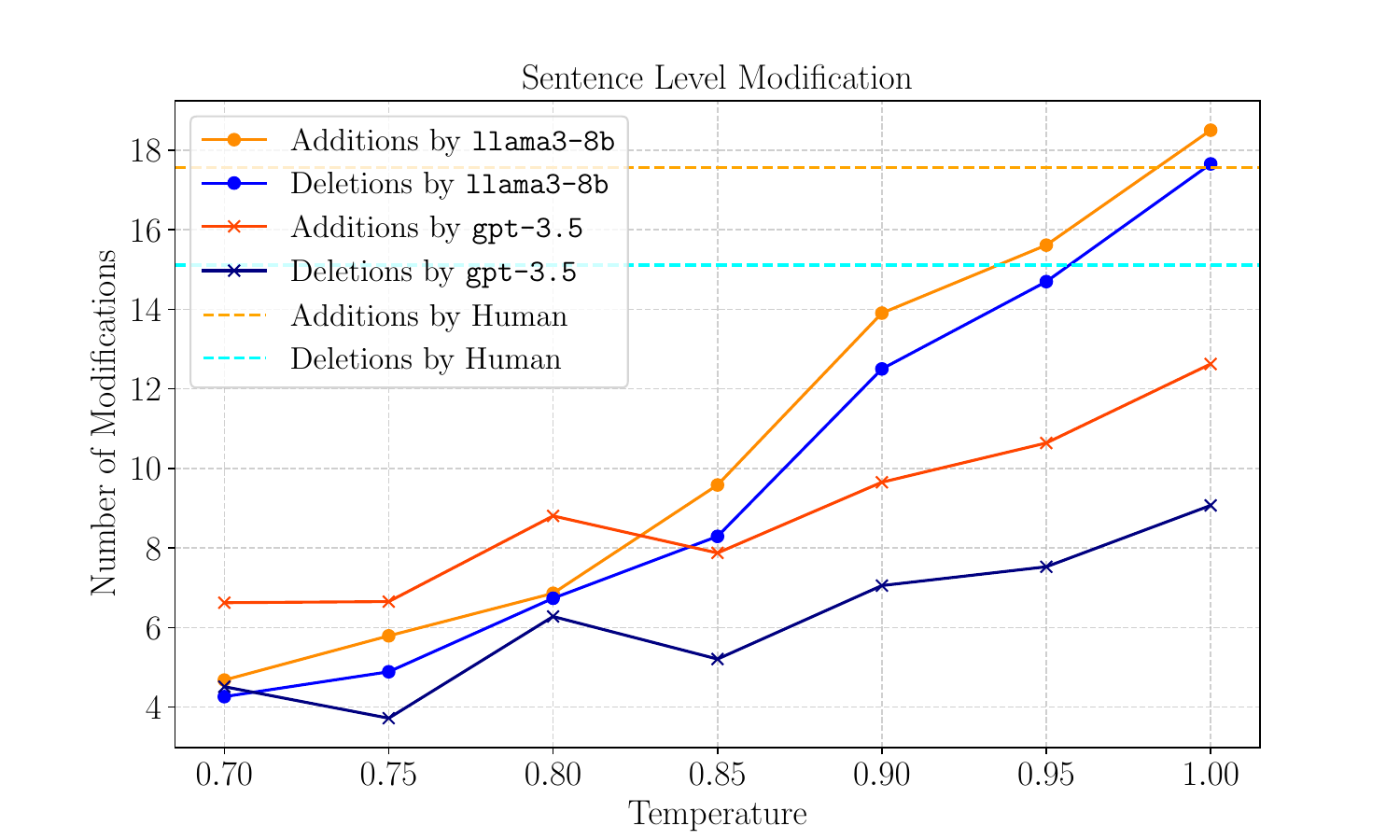}
    \caption{Temperature and sentence-level modifications}
    \label{fig:num_modifications_sentence_level}
\end{figure}

\section{Analysis of Student Simulators}
% \lu{this sec too long, 1p at most}
% \lu{i don't get this para, it seems to just talking about the temperature parameter. intro para should focus on what questions you're asking and answering, e.g. we want to know whether student simulator can faithfully revise based on the feedback, whether our simulator shows similar revision phenomenon as actual student revisions, how to simulate different revision behaviors etc}
% In this section, we study the properties of our trained student simulators and make a comparison with feedback\lu{it's not comparison with feedback, it's actual student revisions based on human feedback} implementation from actual students. 
% One intriguing finding we explore is that the temperature parameter used in auto-regressive decoding can effectively simulate a range of behaviors. This versatility is advantageous as it enables us to check whether a feedback generation system can cater to the diverse array of student personas. In \S \ref{sec:temperature_vs_num_modifications},\lu{also introduce 5.2 and 5.3, or just don't introduce any} we thoroughly evaluate its impact on the number of modifications between the initial version of the essay and the revised version generated by the student simulator.

% \lu{this para can remove}
% \lu{i uncomment this para, check if content is correct and add ref to 4.1 and 4.2}
In this section, we analyze the alignment between the properties of revision in student simulators and actual students, based on peer written feedback. Concretely, we examine the impact of the temperature parameter on the number of modifications (\S\ref{sec:temperature_vs_num_modifications}), revision performance (\S\ref{sec:t_vs_revised_essay_quality}), and faithfulness to feedback (\S\ref{sec:t_vs_faithfulness}), and compare it with the real students' revision process.

% \subsection{Impact of Temperature on the Number of Modifications}
\subsection{Temperature and Revisions}
\label{sec:temperature_vs_num_modifications}
% In this section, our objective is to 
We first analyze the variation in the number of lexical modifications between two student simulators (initialized with \texttt{llama3-8b} and \texttt{gpt-3.5} respectively) using different temperature settings and compare these results to the actual revision process of real students. 
The Ratcliff/Obershelp algorithm~\cite{black2004ratcliff} provides us with a  list of edit operations (additions and deletions) required to transform one sequence into another. We categorize contiguous additions / deletions that involve less than a sentence as word-level modifications, with the number of words involved being counted. For modifications that span an entire sentence or more, we categorize them as sentence-level modifications, with the number of sentences involved being counted.

% \lu{we start with xxx analysis to demonstrate yyy. start each section with a goal sentence, end each section with a takeway and how your later design choice in feedback generator training/test uses this observations}
% For this analysis, we utilized two trained student simulators, initialized with the parameters of \texttt{llama3-8b} and \texttt{gpt-3.5} respectively, to infer the revised writing for each testing datapoint. To measure the lexical difference between the revised and initial writing, we employed the \texttt{difflib} library in \texttt{python}. Similarly, we also compared the actual revised essays implemented by real students with the initial writing. In the \texttt{difflib} library, the lexical difference between two textual sequences is calculated using the Ratcliff/Obershelp algorithm~\cite{black2004ratcliff}. This algorithm provides a list of edit operations (additions and deletions) required to transform one sequence into another.\lu{shorten, no need for this much detail} 

% We categorized the changes into two types based on the number of contiguous words added or deleted: word-level modifications and sentence-level modifications.\lu{this is important detail, what is counted as word vs sentence? anything is not a full sentence will be word? are these two types exclusive?}

% \lu{this paragraph is very long, though they're useful observations, we should only keep the ones that are important for sys design choice explanation, or linked to later explanation on feedback quality}
From the plots shown in Figures \ref{fig:num_modifications_sentence_level} and \ref{fig:num_modifications_word_level}, we see that the number of elements (words/sentences) added or deleted by the student simulators increases as the temperature is increased. This observation aligns with expectations, as higher temperature settings introduce more randomness during decoding, leading to increased number of alterations~\cite{renze2024effect}.
% Consequently, the models are encouraged to be creative and make more substantial alterations to the content~\cite{renze2024effect}. 
% As students are expected to write essays whose number of words lie in a narrow band ($400$-$500$ words), we observe that the number of additions and deletions are approximately equal. 
Furthermore, the \texttt{gpt-3.5}-based student simulator tends to be more conservative than the \texttt{llama3-8b}-based counterpart, with a greater inclination towards preserving the original content. 
% This implies that the former model exhibits a greater inclination towards preserving the original content, while the latter demonstrates a slightly more liberal approach by introducing additional modifications. 
% Finally, we observe that the number of modifications from real students aligns with student simulators' outputs at higher temperature settings.

% \subsection{Impact of Temperature on the Revised Essay Quality}
\subsection{Temperature and Revision Quality}
\label{sec:t_vs_revised_essay_quality}

\begin{figure}[t]
    \centering
    \includegraphics[scale=0.3]{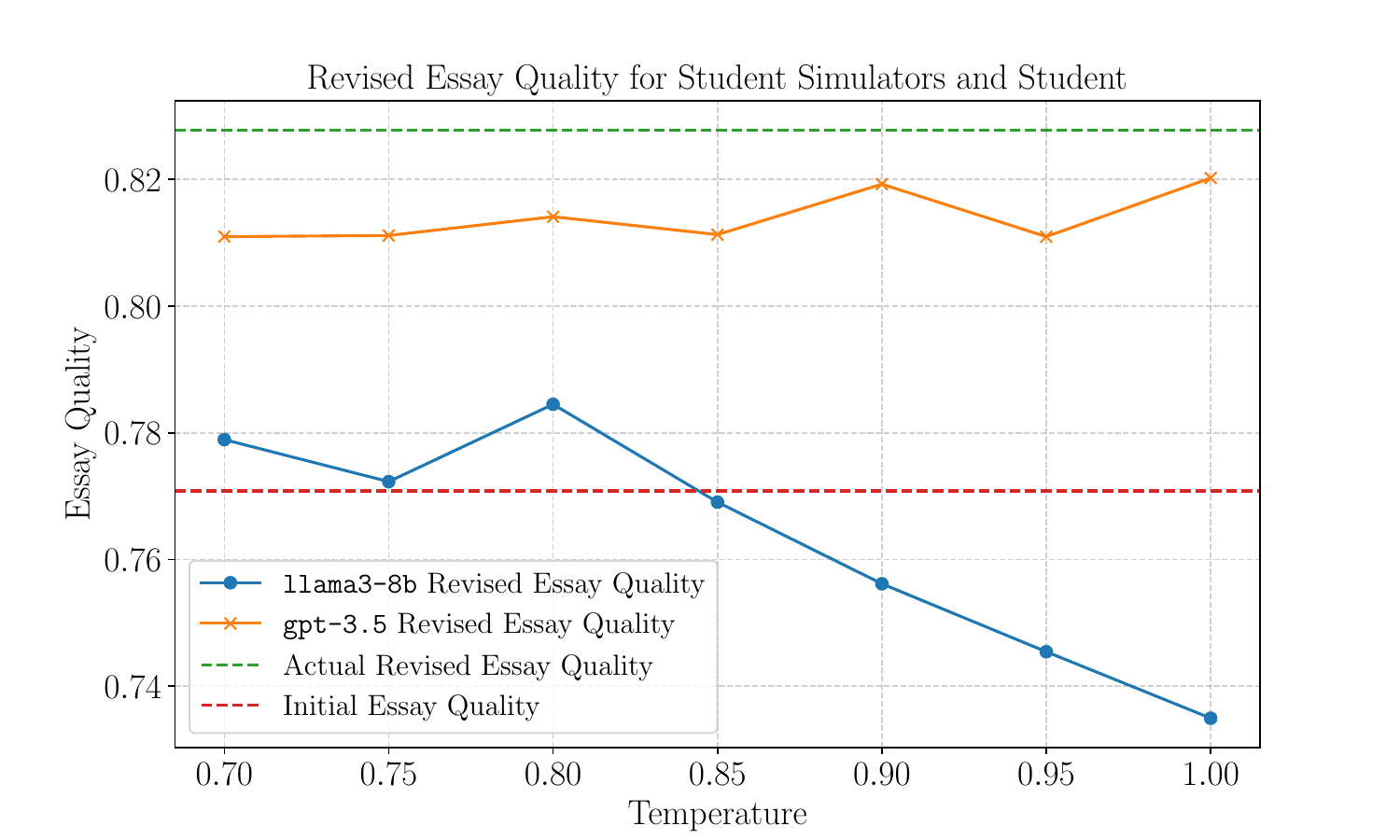}
    \caption{The quality of the revised essay by student simulators vs. actual students.}
    \label{fig:revised_essay_quality}
\end{figure}

In this analysis, we compute the quality of the revised essays based on peer feedback from both student simulators and actual students using \texttt{gpt-4} as the judge. To validate the use of \texttt{gpt-4} in evaluating essay quality, we compared its inferred scores with the scores assigned by teaching instructors of the Economics course to actual students' final revised essays. The mean squared error was $0.082$ after normalizing the scores between $0$ and $1$, suggesting that the inferred scores are reliable and closely align with the expert-assigned scores.

Based on the findings presented in Figure \ref{fig:revised_essay_quality}, it is evident that actual students exhibit a higher level of effectiveness in incorporating feedback compared to the student simulators. This is a desirable outcome as it creates a more challenging environment for our feedback generator during training and testing with student simulators. Among the student simulators, the model based on \texttt{gpt-3.5} demonstrates superior implementation performance, with a slight improvement in the quality of revised essays as the temperature increases and reasonable alignment with the quality of revision from real students. \textit{This makes \texttt{gpt-3.5} a suitable approach for automatically assessing feedback effectiveness.} 
On the other hand, the \texttt{llama3-8b} based student simulator demonstrates a modest enhancement in quality at lower temperatures but experiences a decline beyond a temperature of $0.8$.
%Higher temperatures encourage more content changes, which can adversely affect essay quality.

\subsection{Temperature and Revision Faithfulness}
\label{sec:t_vs_faithfulness}

\begin{table}[t]
    \tiny
    \centering
    \setlength\tabcolsep{3.9pt}
    \begin{tabular}{c|ccc|ccc|c}
        \midrule
        \multirow{3}{*}{\textsc{Approach}} & \multicolumn{6}{c|}{\textsc{Student Simulators}} & \textsc{Real} \\
        & \multicolumn{3}{c|}{\texttt{llama3-8b}} & \multicolumn{3}{c|}{\texttt{gpt-35-turbo}} & \textsc{Students} \\
        & $0.7$ & $0.85$ & $1.0$ & $0.7$ & $0.85$ & $1.0$ & \\
        \midrule
        \# of Faithful Rev. ($F$) & \multirow{1}{*}{1.1} & \multirow{1}{*}{2.4} & \multirow{1}{*}{3.6} & \multirow{1}{*}{1.8} & \multirow{1}{*}{1.6} & \multirow{1}{*}{1.9} & \multirow{1}{*}{4.5}\\
        % Revisions ($F$) & & & & & & & \\
         % & & & & & & & \\
        \midrule
        \# of Unfaithful Rev. ($U$) & \multirow{1}{*}{1.4} & \multirow{1}{*}{1.3} & \multirow{1}{*}{4.1} & \multirow{1}{*}{0.5} & \multirow{1}{*}{0.8} & \multirow{1}{*}{2.0} & \multirow{1}{*}{1.4}\\
        % Revisions ($U$) & & & & & & & \\
         % & & & & & & & \\
        \midrule
        % $\gamma = \frac{F}{U}$ & 1.2 & \textbf{3.3} & 1.1 & \textbf{2.5} & 2.2 & 1.2 & 17.7\\
        $\gamma = \log\left(\frac{F}{U}\right)$ & -0.1 & \textbf{0.3} & -0.1 & \textbf{0.6} & 0.3 & -0.1 & 0.5\\
        \bottomrule

    \end{tabular}
    \caption{The average number of faithful and unfaithful revisions from student simulators operating at different temperatures vs. actual student revisions. $\gamma$ assigns higher value to faithful revisions that adhere to the feedback without making additional content changes.}
    \label{tab:t_vs_faithfulness}
\end{table}

Next, we assess how faithful the student simulators are in implementing the feedback, in comparison to the revisions made by real students. For this purpose, we broke down the feedback into a list of distinct recommended changes, and then examined how many of these changes were incorporated in the revised writing, as compared to the initial version. We categorized the changes as either \textbf{faithful} (i.e., they adhered to the provided feedback) or \textbf{unfaithful} (i.e., they went beyond the scope of the feedback and made additional changes).
% Since real students may consider the initial content and feedback and make additional changes beyond the suggested revisions, we also count the number of \textbf{unfaithful} revisions for the student simulators and the real student. Conversely, the \textbf{faithful} revisions refers to the changes that respects the provided feedback. 
For $10$ samples, the number of these instances at $3$ temperature values for the student simulators was manually annotated by a fluent English speaker. This resulted in the analysis of $(3 + 3 + 1) \times 10 = 70$ revised essays, with $3$ revisions from each of the student simulators and one actual revision.

Comparing the revisions generated by student simulators with those made by humans, we find that the simulators produce more unfaithful revisions and fewer faithful revisions, as shown in Table \ref{tab:t_vs_faithfulness}. While previous experiments indicate that higher temperatures result in more content-level modifications from the simulators, they do not always align with the provided feedback. To quantify the faithfulness of modifications, we define $\gamma$ as the logarithm of the ratio between the number of faithful and unfaithful modifications. These results suggest that there is still room for improvement in the faithfulness of simulated revisions compared to actual human revisions.

% \lu{this para can remove}
% \textbf{Takeaways}: The above experiments showcase the ability to simulate diverse behaviors by adjusting the temperature and manipulating their properties like the number of modifications, revision quality, and faithfulness. To further explore the impact of training a feedback generator on these behaviors and evaluate its performance across various personas, we examine the effectiveness of our feedback generators trained using the student simulator operated at temperatures: $0.7, 0.85, 1.0$. We also test their effectiveness under the same temperature settings.

% \section{Feedback Generator: Implementation Details}
\section{Feedback Generator Setups}
% \lu{if this sec is about feedback generator implementation details, either put it in next sec, or add `feedback generation' in the title, otherwise it's confusing. alternatively put it as the last concluding design choice paragraph in sec 4}
For our iterative optimization approach, \textbf{we use \texttt{llama3-8b} based student simulator for training and emphasize on \texttt{gpt-3.5} based one for testing}.  
We do this for the following three reasons:  
\textbf{(1)} \texttt{gpt-3.5} based student simulator better aligns with actual student revision, as demonstrated in \S \ref{sec:t_vs_revised_essay_quality} and thus provides a more realistic testing environment to measure implementation performance. 
\textbf{(2)} Using \texttt{gpt-3.5} based student simulator is prohibitively expensive when used in conjunction with an iterative optimization approach. Using \texttt{llama3-8b} as a student simulator makes our research more accessible due to lower training cost and easy access to the open-source models.
\textbf{(3)} Training and testing on the same student simulator would not provide conclusive evidence of the effectiveness of our approach, as it might learn to exploit one type of student simulator while performing poorly on others.
For completeness, however, we also include the effectiveness of the generated feedback using the \texttt{llama3-8b} based student simulator.

Refer Appendix \ref{sec:hyperparameter_settings} for more details.
% contains more information regarding the hyperparameters employed during the training of the student simulator, initialization of the feedback generator, and its iterative optimization.

\section{Results for Feedback Generation}

We consider the following types of baselines: (1) \textbf{\texttt{gpt-3.5 / gpt-4}}: Using enterprise LLMs as a few-shot feedback generator by sampling in-context examples from peer-written feedback from the train-split. (2) \textbf{\texttt{sft-from-human}}: Fine-tuning \texttt{llama3-8b} on peer review feedback.
% \begin{enumerate}
%     \item \textbf{\texttt{gpt-3.5 / gpt-4}}: We use enterprise LLMs as a few-shot feedback generator by sampling in-context examples from peer-written feedback from the train-split.
%     \item \textbf{\texttt{sft-from-human}}: We fine-tune \texttt{llama3-8b} on peer review feedback. %{Note that this approach, along with the following ones, has significantly fewer parameters compared to the above few-shot prompting approach.}
%     % \item \textbf{\texttt{sft-from-gpt4}}: The model initialized with \texttt{llama3-8b} parameters is fine-tuned over \texttt{gpt-4} generated feedback for the train split. 
%     % (this will be added to the appendix)
% \end{enumerate}

Our method variants are named as ``\textbf{\model, $\tau=x$}" where the \texttt{llama3-8b} based feedback generator is initialized with \texttt{sft-from-human} and trained using the iterative optimization framework described in \S \ref{sec:model} along with the \texttt{llama3-8b} based student simulator executed at temperature $x$. 
% We initialize our feedback generator from \texttt{sft-from-human} to show that our approach relies minimally on supervision from \texttt{gpt-4}, and solely uses it for evaluating the quality of revised essays generated by the student simulator.\lu{can remove this sent}

\subsection{Intrinsic Evaluation}
\label{sec:intrinsic_evaluation}
\begin{table}[t]
\setlength\tabcolsep{5pt}
\scriptsize
    \centering
    \begin{tabular}{c|cccc|c}
        \midrule
        \multirow{2}{*}{\textsc{Approach}} & \multicolumn{4}{c|}{\textsc{Pedagogical Evaluation}} & \multirow{2}{*}{\textsc{Avg.}}\\
        & \sc RGQ & \sc EAL & \sc DM & \sc MSSC &  \\
        \midrule
        \texttt{gpt-3.5} & 70.6 & \best{80.0} & \secondbest{78.6} & \best{60.0} &  72.3\\
        \texttt{gpt-4} & \secondbest{71.4} & \best{80.0} & \best{79.2} & 59.4 &  \best{72.5}\\
        \texttt{sft-from-human} & 65.8 & 67.8 & 65.6 & 53.3 & 63.1\\
        \midrule
        \multicolumn{6}{c}{\model, $\tau = 0.7$} \\
        \midrule
        \textit{Iteration 1} & 66.7 & 77.2 & 76.4 & 57.2 & 69.4 \\
        \textit{Iteration 2} & 68.6 & 79.2 & 77.8 & 59.2 & 71.2 \\
        \textit{Iteration 3} & 70.6 & \secondbest{79.4} & 78.3 & \secondbest{59.7} & 72.0\\
        \midrule
        \multicolumn{6}{c}{\textsc{\model, $\tau = 0.85$}} \\
        \midrule
        \textit{Iteration 1} & 70.8 & 78.3 & 75.0 & 58.3 & 70.6\\
        \textit{Iteration 2} & 70.6 & 79.2 & 77.8 & 58.6 & 71.6 \\
        \textit{Iteration 3} & \best{71.9} & 79.2 & \best{79.2} & 59.2 & \secondbest{72.4}\\
        \midrule
        \multicolumn{6}{c}{\textsc{\model, $\tau = 1.0$}} \\
        \midrule
        \textit{Iteration 1} & 62.2 & 62.5 & 61.4 & 53.0 & 59.8\\
        \textit{Iteration 2} & 70.8 & 73.6 & 69.4 & 57.2  & 67.8\\
        \textit{Iteration 3} & 70.8 & 76.9 & 75.8 & 58.0 & 70.4 \\
        \bottomrule
    \end{tabular}
    \caption{
    % \lu{this table can be in one column since there are only 6 numbers per row (now so much white space), you can use Iter. 1, gpt-3.5 for short names. respects guided questions can be GQ, then in the caption can say `GQ: respects Guided Questions', the rest are the same.}
    % \lu{since the bests are pretty much gpt, i suggest we color the `best' per column and our best model variant that's better than sft-from-human if this looks better and can deliver more consistent messages.}
    We evaluate the intrinsic quality of the generated feedback in terms of pedagogical alignment. \best{Green} and \secondbest{Blue} represents best and second-best performance respectively. 
    % Despite not being explicitly trained for the task, our approach demonstrates significant improvement in pedagogical aspects as compared to \texttt{sft-from-human} which was used to initialize our feedback generator for each temperature.
    }
    \label{tab:pedagogical_evaluation}
\end{table}

To intrinsically evaluate the quality of the feedback generated from different approaches, we employ LM as the judge~\cite{chevalier2024language,ke2023critiquellm} and evaluate along the following \textbf{four major pedagogical dimensions}~\cite{deepmindeducation}: 

\begin{itemize}
    \item \textbf{Respects Guided Questions (RGQ)}: Given that the assignment uses scripted feedback, the generated responses are expected to follow a template with $6$ prompts, each accompanied by a targeted feedback. Here, we assess how well each feedback component aligns with its respective prompt. To view the prompt template, refer to Appendix \ref{sec:peer_review_prompt}.
    
    \item \textbf{Encourages Active Learning (EAL)}: Measures how well the feedback guides the students to make improvements on their own without explicitly revealing the concrete changes. 
% For example, providing feedback such as "{\it The author could utilize quantitative analysis to elucidate the effects of a minimum wage increase and a policy banning automation}" does not directly reveal explicit changes and guide students towards areas of improvement.\lu{may need to remove this sent to save space}
    \item \textbf{Deepens Metacognition (DM)}: Determine the effectiveness of the feedback in identifying and addressing student errors and misconceptions within the essay. 
% It is important to note that this assessment focuses on how well the feedback identifies and offers suggestions, rather than solely evaluating its intrinsic correctness.
% \lu{this is still too abstract, is this for correctness? if so, make it clear since readers will care about the correctness of feedback}
    \item \textbf{Motivates and Stimulates Student Curiosity (MSSC)}: Assess how well the feedback maintains a positive and encouraging tone that fosters curiosity and motivation. 
\end{itemize}

% \textbf{(5) Adapts to Essay Quality(AEQ)}: Evaluate the feedback's ability to adjust based on the quality of the essay.\lu{this one isn't clear, specific/customized to the essay?}

% \lu{we need to add some discussion on whether gpt judge can do this job well. if we don't have quantatitive analysis, at least add discussion on human inspections are used for prompt engineering to ensure they work reasonably well, though we should still be prepared that reviewers will attack us on this eval since no human eval is conducted whatsoever}
For the score generated by \texttt{gpt-4} in relation to a sample feedback, please see Appendix \ref{sec:example_pedagogical}. 
% The performance of \texttt{gpt-4} as an evaluator for pedagogical alignment has been studied by \citet{deepmindeducation}, who discovered significant alignment between the scores generated by \texttt{gpt-4} and those produced by human evaluators along pedagogical dimensions.
The idea of using a critic LLM to automatically assess the quality of feedback based on pedagogical aspects was inspired by \citet{deepmindeducation}, who demonstrated a strong correlation between the generated scores and those provided by humans. In our study, we utilized \texttt{gpt-4} to automatically assign pedagogical scores to the feedback. 

% \lu{add a paragraph title for this gpt4 vs human analyais. move this paragraph after the results, since here you didn't say the rating is from 0-5, not 1-5? is this what human rates too?} 
% The inter-annotator correlation scores for "respects guided questions", "encourages active learning", "deepens metacognition" and "motivates and stimulates student curiosity" are 0.27, 0.54, 0.64 and 0.71

% \lu{give a title here too}
% \lu{this part is long, either create several paragraphs, or highlight key takeaways using italics or bold}
\paragraph{Results from \texttt{gpt-4} based pedagogical evaluation.}
We used \texttt{gpt-4} to assign scores ranging from 0 to 5 for each metric, representing lowest to highest quality. The average quality for each metric was then calculated, normalized between 0 and 100, and presented in Table \ref{tab:pedagogical_evaluation}. \textit{Our approach significantly improves performance for \texttt{sft-from-human} and achieves comparable results to enterprise LLMs, despite having significantly fewer parameters.} This demonstrates the effectiveness of our approach without requiring high-quality feedback.

Among our models trained with student simulators at different temperatures, we observe the most significant improvements in the model trained at a temperature of $0.85$. 
We believe that at lower temperatures, the student simulators only incorporate a limited number of feedback elements, as discussed in \S \ref{sec:t_vs_faithfulness}. This limitation prevents the appropriate selection of desirable and undesirable feedback at each stage of preference learning. Conversely, at higher temperatures, the student simulator demonstrates a higher degree of unfaithfulness, leading to poor implementation performance, as demonstrated in \S \ref{sec:t_vs_faithfulness} and \S \ref{sec:t_vs_revised_essay_quality}, respectively. By using \texttt{sft-from-human} as the foundation for our feedback generator across various temperature settings, we achieve a notably improved model in terms of pedagogical alignment.
% \lu{can remove} 
% Moreover, our model requires fewer parameters and incurs lower compute costs, while still delivering comparable performance to \texttt{gpt-3.5/gpt-4}.

\paragraph{Validation of \texttt{gpt-4}'s pedagogical evaluation.} 
To validate the pedagogical evaluation using \texttt{gpt-4}, we used 21 pairs of essays and peer reviews and had two proficient English annotators assess the peer reviews across various pedagogical dimensions by providing a score between 0 and 5 for each metric. We then calculated the Pearson correlation between the average normalized scores assigned by humans and those inferred by \texttt{gpt-4}. 

We noted a moderate-to-high correlation between the generated and averaged annotated scores for dimensions such as "respects guided questions", "encourages active learning", and "deepens metacognition", with respective values of 0.31, 0.40, and 0.71. However, the dimension "motivates and stimulates student curiosity" demonstrated a lower correlation of 0.20, likely due to the subjective nature of assessing positivity and the encouraging tone of feedback. In terms of average pedagogical score, we observed a correlation of 0.40. These measures of correlation justify the validity of pedagogical evaluation from \texttt{gpt-4}.

\subsection{Extrinsic Evaluation}
\label{sec:extrinsic_evaluation}
\begin{table}[t]
\scriptsize
    \setlength\tabcolsep{8pt}
    \centering
    \begin{tabular}{c|ccc|c}
        \midrule
        \multirow{2}{*}{\textsc{Approach}} & \multicolumn{3}{c|}{\textsc{Student Simulator}} & \multirow{2}{*}{\textsc{Avg.}}\\
        & 0.7 & 0.85 & 1.0 \\
        \midrule
        \texttt{gpt-3.5} & 76.3 & 77.1 & 76.9 & 76.8\\
        \texttt{gpt-4} & 76.6 & 77.0 & 77.4 & 77.0\\
        \texttt{sft-from-human} & 78.9 & 78.8 & 80.3 & 79.4\\
        \midrule
        \multicolumn{5}{c}{\model, $\tau = 0.7$} \\
        \midrule
        \textit{Iteration 1} & \best{80.1} &	\secondbest{79.9} & 79.7 & \secondbest{79.9}\\
        \textit{Iteration 2} & 77.1 &	79.4 &	80.0 & 78.9\\
        \textit{Iteration 3} & 79.0 & 	77.5 & \best{80.9} & 79.1 \\
        \midrule
        \multicolumn{5}{c}{\model, $\tau = 0.85$} \\
        \midrule
        \textit{Iteration 1} &	79.3 &	\best{80.0} &	\secondbest{80.8} & \best{80.0} \\
        \textit{Iteration 2} & 	79.0 & 	74.8 &	78.7 & 77.5\\
        \textit{Iteration 3} &	79.1 & 	79.5 &	79.5 & 79.4\\
        \midrule
        \multicolumn{5}{c}{\model, $\tau = 1.0$} \\
        \midrule
        \textit{Iteration 1} & 	79.2 &	77.2 & 	77.3 & 77.9\\
        \textit{Iteration 2} & 	\secondbest{79.7} & 	\best{80.0} &	78.2 & 79.3\\
        \textit{Iteration 3} & 	78.0 &	78.9 &	76.6 &  77.8\\
        \bottomrule
        
    \end{tabular}
    \caption{
    % \lu{with llama simulator result, colors are everywhere, can we just report gpt results here and have llama results in appendix?}
    Extrinsic evaluation using \texttt{gpt-3.5} based student simulator. \best{Green} and \secondbest{Blue} represents best and second-best performance respectively. Each experiment was repeated for $5$ different seeds to mitigate the impact of randomness.}
    \label{tab:extrinsic_evaluation}
\end{table}

% \lu{highlight takeaways in this sec too}
Next, we use trained student simulators to gauge the efficacy of feedback generated by different systems. Although we discuss results by using \texttt{gpt-3.5} based student simulator, we also include the results computed using \texttt{llama3-8b} based student simulator in Table \ref{tab:extrinsic_evaluation_llama} of Appendix \ref{sec:extrinsic_evaluation_llama}. \texttt{gpt-3.5}-based simulator aligns more closely with actual human performance and was not employed in training our models, making the effectiveness evaluation more reliable and trustworthy.

% For each of the approaches, we employed greedy decoding to generate feedback for each instance in the testing dataset. Subsequently, the student simulators were executed at three different temperatures: 0.7, 0.85, and 1.0, resulting in revised essays. Since temperature sampling introduces randomness in the sampled results, we generated five different essays at each temperature. This approach allows us to average the quality of the generated essays and mitigate the effect of randomness. Finally, we use the course rubric prompt (refer to Appendix \ref{sec:grade_essay_prompt}) to compute the score. Since this score is based on multiple aspects, we calculated the average score across different aspects and normalized it between $0$ and $100$.

For each approach, we use greedy decoding to generate feedback. The student simulators are executed at $3$ different temperatures: $0.7$, $0.85$, and $1.0$, with $5$ different seeds to mitigate the impact of randomness. Finally, we compute a score based on the course rubric prompt (refer to Appendix \ref{sec:grade_essay_prompt}), averaging the scores across different aspects and normalizing it between $0$ and $100$.

\textit{Table \ref{tab:extrinsic_evaluation} demonstrates that our approaches consistently outperform enterprise LLMs like \texttt{gpt-3.5/gpt-4}}. Interestingly, the results also indicate a lack of correlation between extrinsic and intrinsic evaluations. While \texttt{sft-from-human} exhibits better extrinsic performance, its intrinsic performance is lower compared to the few-shot approaches. One possible explanation for this discrepancy is that \texttt{sft-from-human} provides explicit feedback, which may negatively impact the "Encourages Active Learning" metric (Table \ref{tab:pedagogical_evaluation}), but contribute to a more effective revision process. In most cases, one round of iterative optimization leads to the best extrinsic performance. However, it is important to note that these findings are specific to the experiments conducted on a domain-specific course and may not necessarily apply to broader datasets or different contexts.

\subsection{Additional Analyses}
\label{sec:additional_analyses}

% \lu{didn't read onwards, too long, need to significantly shrink}
\noindent \textbf{RQ1: How does our training algorithm influence different feedback categories with the number of refinement iterations?} 
In this analysis, we break down the feedback into distinct components and classify them into one of $3$ categories: \textbf{praise}, \textbf{solution}, or \textbf{problem}. Our primary objective is to verify whether \model effectively adjusts the frequency of these elements in a manner that is consistent with the learning sciences research on maximizing feedback effectiveness~\cite{lizzio2008feedback,nelson2009nature,cho2010student}.

\begin{figure}[t]  
    \centering  
    \begin{subfigure}{0.3\textwidth}  
        \includegraphics[width=\linewidth]{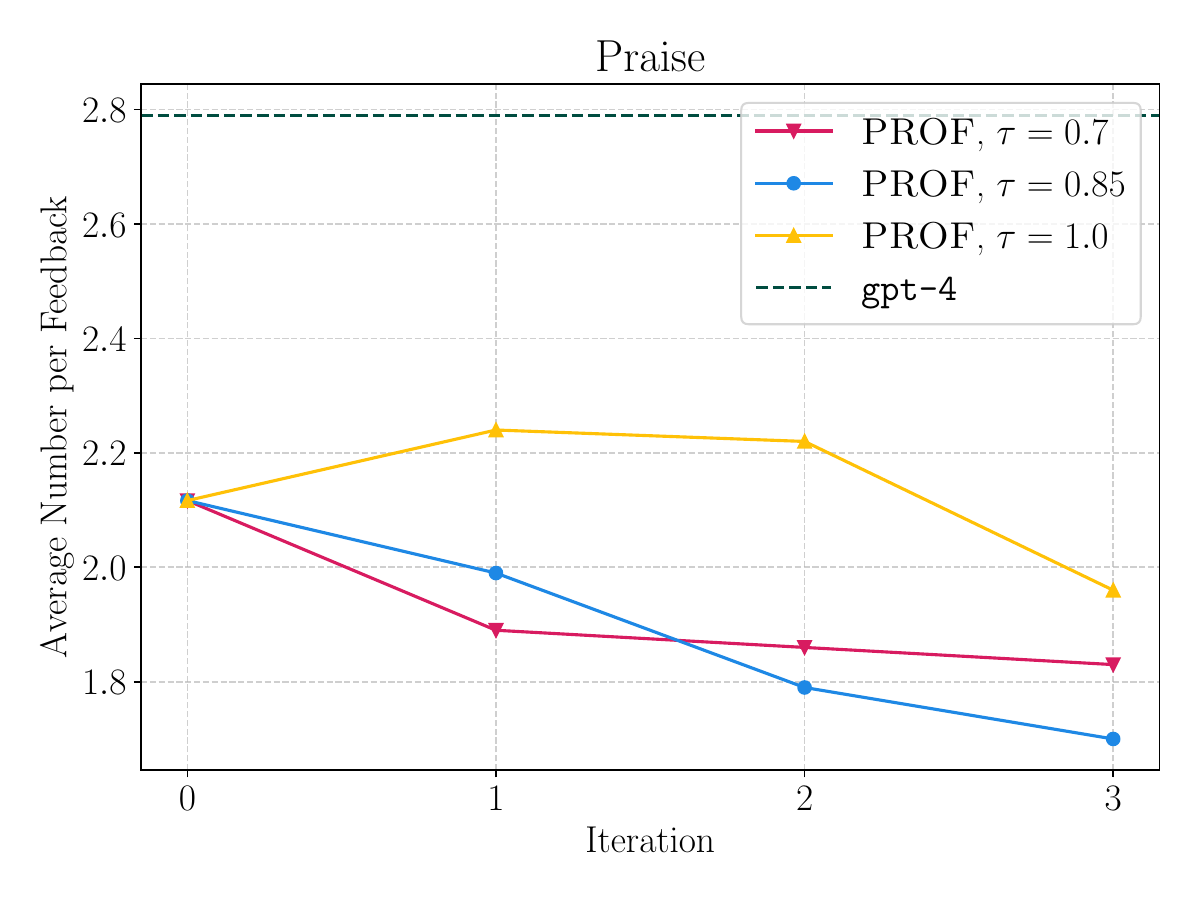}  
        \caption{Number of \textbf{Praise} segments}  
        \label{fig:evolution_praise}  
    \end{subfigure}  
    \begin{subfigure}{0.3\textwidth}  
        \includegraphics[width=\linewidth]{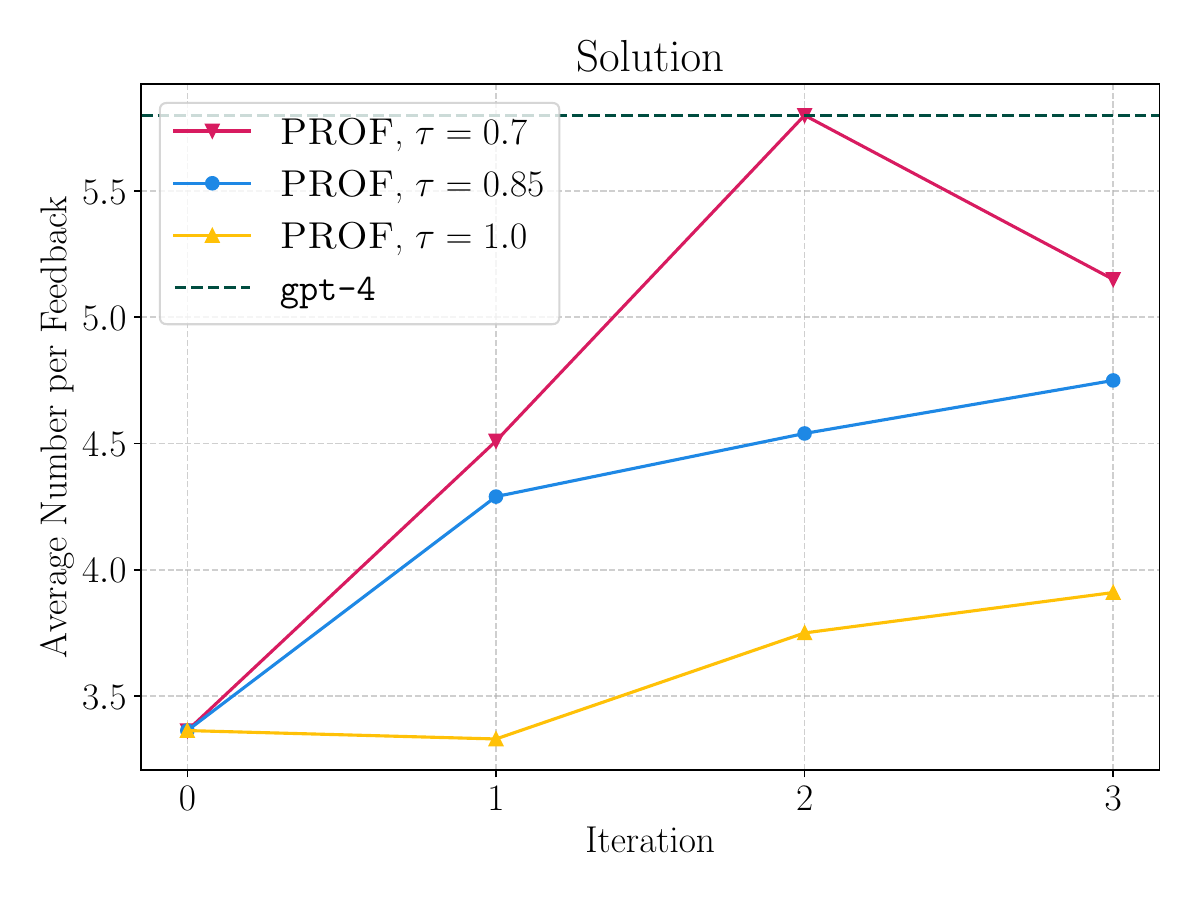}  
        \caption{Number of \textbf{Solution} segments}  
        \label{fig:evolution_solution}  
    \end{subfigure}
    \begin{subfigure}{0.3\textwidth}  
        \includegraphics[width=\linewidth]{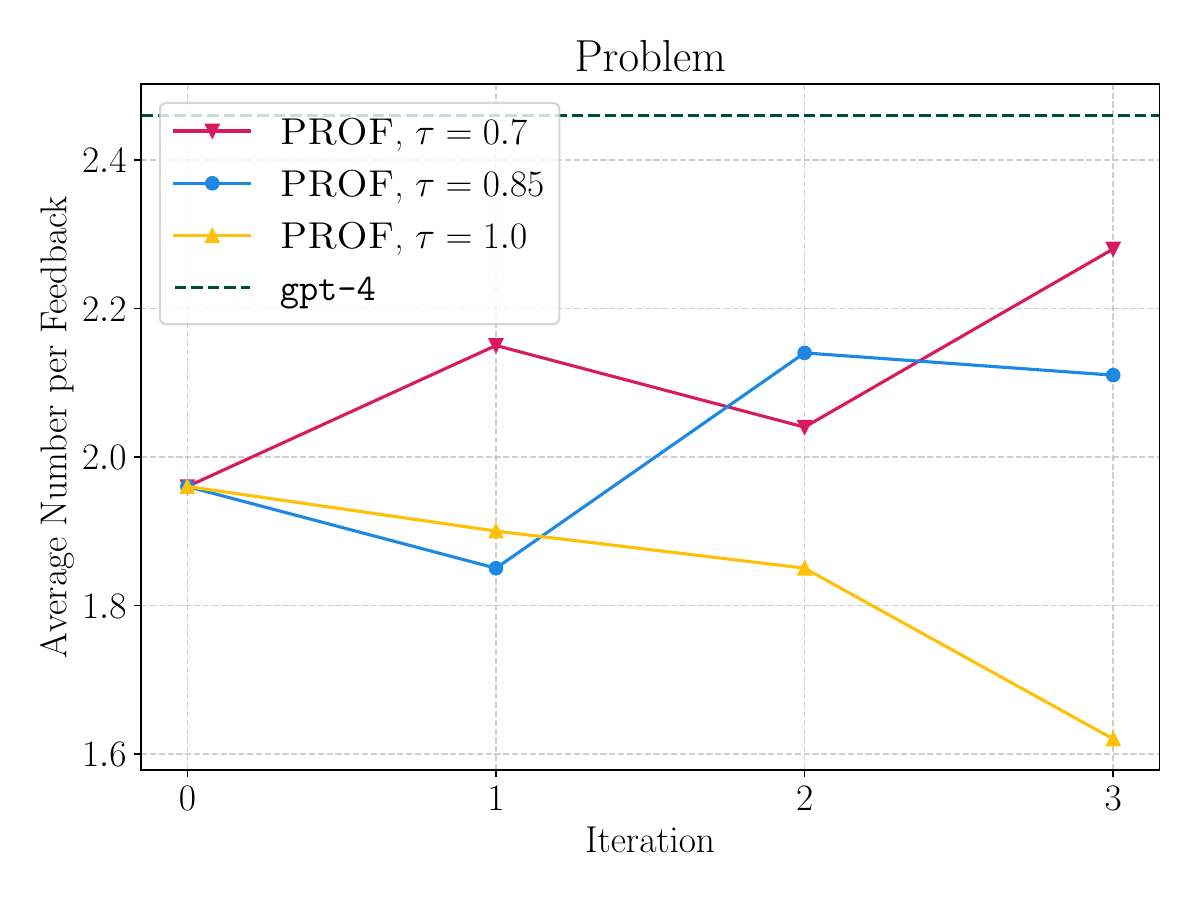}  
        \caption{Number of \textbf{Problem} segments}  
        \label{fig:evolution_problem}  
    \end{subfigure}    
    \caption{Evolution of segments}  
    \label{fig:evolution}  
\end{figure}

% In this analysis, we break down the feedback into distinct components, ensuring that each component encapsulates a coherent change. These components can range from a single sentence to several sentences. We then classify each segment into one of three categories: \textbf{praise}, \textbf{problem}, or \textbf{solution} as these categories have undergone extensive research in learning science to understand their impact on student revision performance~\cite{lizzio2008feedback,nelson2009nature,cho2010student}. Our objective is to verify whether \model effectively adjusts the frequency of these elements to enhance student implementation performance, aligning with learning science research~\cite{ferris1997influence,sugita2006impact}. 
% \lu{say since they are shown by education sciences or learning sciences as important to .... then cite. we need to make more connections with literature to justify why we're doing this} 
%
If a segment solely describes an issue, it is labeled as a \textbf{problem}. If segment contains both problem and solution, it is still categorized as a \textbf{solution}. To view how our algorithm influences the number of \textbf{praise} elements, refer to 
 Figure \ref{fig:evolution_praise}. We notice that the average number of \textbf{praise} elements decreases with more optimization steps, which is corroborated by previous research indicating that praise has minimal impact on student performance~\cite{kluger1996effects,ferris1997influence}.
% This categorization is based on the understanding that the presence of a solution in a segment provides explicit guidance to students during content revision, unlike problem segments which do not offer explicit guidance~\cite{nelson2009nature}. 
% \lu{can you rewrite this to if a sentence contains both xx and yy, we label it as yy because....}
 
 In general, the average number of \textbf{solution} and \textbf{problem} elements in the feedback (refer Figure \ref{fig:evolution_solution} and \ref{fig:evolution_problem} respectively) increases with the number of iterations which is known to impact implementation performance as supported by many prior works~\cite{hayes1987cognitive,matsumura2002teacher,bitchener2005effect,sugita2006impact}.
 % \lu{can refer to the figs here? is this para still needed? since later you mentioned the temp at 1.0 again? check for redundancy}

Among our models trained at different temperatures, \model, $\tau = 0.7$ significantly increases the average number of \textbf{solution} elements compared to the model trained at a temperature of $1.0$ as shown in Figure \ref{fig:evolution_solution}. When training with the student simulator at a high temperature ($1.0$), the feedback generator is guided to minimally increase the number of solution segments. This is intuitive because the student simulator can introduce numerous changes that may lower the quality of the essay at higher temperatures. If there are too many solution elements, the model may make excessive changes that could degrade the essay's quality.
% Providing many solution segments may have a negative impact on the quality of the essay as the student simulator, at higher temperatures, implements numerous changes that can potentially deteriorate the overall quality of the essay.\lu{this sentence isn't grammatical, break it down to 2 sents?}
% Considering that the student simulator makes numerous changes at $1.0$ temperature that detriorates the quality of the 
% Recall that our student simulator at temperature $1.0$ was highly liberal in revising content, leading our feedback generator to generate feedback with only a modest increase in the number of solutions for best effectiveness. Increasing the number of solution components too much can cause the student simulator to make substantial changes to the essay content at higher temperature and as the \texttt{llama3-8b} based student simulator is ineffective when making numerous changes at higher temperatures, the optimization process results in only a modest increase in the number of \textbf{solution} elements from the feedback generator. 
Conversely, with the student simulator being very conservative at a temperature of $0.7$, our optimization algorithm responds by more aggressively increasing the number of solution components.

While \textbf{solution} provides explicit guidance on improving the initial writing, the lack of this guidance from the \textbf{problem} elements coupled with the ineffectiveness of our \texttt{llama3-8b}-based student simulator at higher temperatures ($1.0$) make the feedback generator produce fewer \textbf{problem} elements as the number of the training iterations increases (refer Figure \ref{fig:evolution_problem}).
% \lu{potential change; fig should be placed after solution since it's introduced later} 
% For lower temperatures, as the student simulator is more effective and conservative in revising the content, the feedback generator is guided towards generating more \textbf{problem} elements with the number of iterations. 
% {\it To increase the frequency of both problem and solution segments, we recommend running the \model at a lower $\tau$. However, if explicit guidance is required for a specific use-case, executing the \model at a higher $\tau$ is advisable.}

% \lu{ok, there's lots of descriptions on `what's going on' in the figure, but what's the takeaway message? is our model doing better? should ppl use high/low temperature for training or test for what purpose? i stopped reading here, i thinh this sec needs some polishing to be clearer}

% \textbf{Further categorization of Problem / Solution Feedback elements with the number of Refinement iterations}:
As the \textbf{problem} and \textbf{solution} feedback elements are influential in improving feedback effectiveness, we further sub-categorize these elements and analyze how they evolve with the number of refinement iterations. 
% Our aim is to explore the capability of our training algorithm in guiding the feedback generator to make appropriate decisions for the specific persona it has interacted with.

\noindent \textbf{RQ2: How does the faction of problem or solution feedback segments associated with a local Scope vary with the number of refinement iterations?} The feedback scope is broadly categorized into two classes: (a) {\bf local scope}, which focuses on specific words, phrases, sentences, or paragraphs and is associated with narrow aspects such as surface features; (b) {\bf global scope}, involves considering multiple parts or the entirety of the writing. Both local-scope and global-scope feedback have been observed to result in improved essay quality after implementation~\cite{olson1987effects,lin2001web,miller2003effect}. To ensure effective feedback for enhancing essay quality, it is important to include an appropriate proportion of both locally and globally scoped problem and solution segments, rather than focusing solely on one type.

 \begin{figure}[t]
\centering
\begin{subfigure}[b]{0.33\textwidth}
\centering
\includegraphics[width=\textwidth]{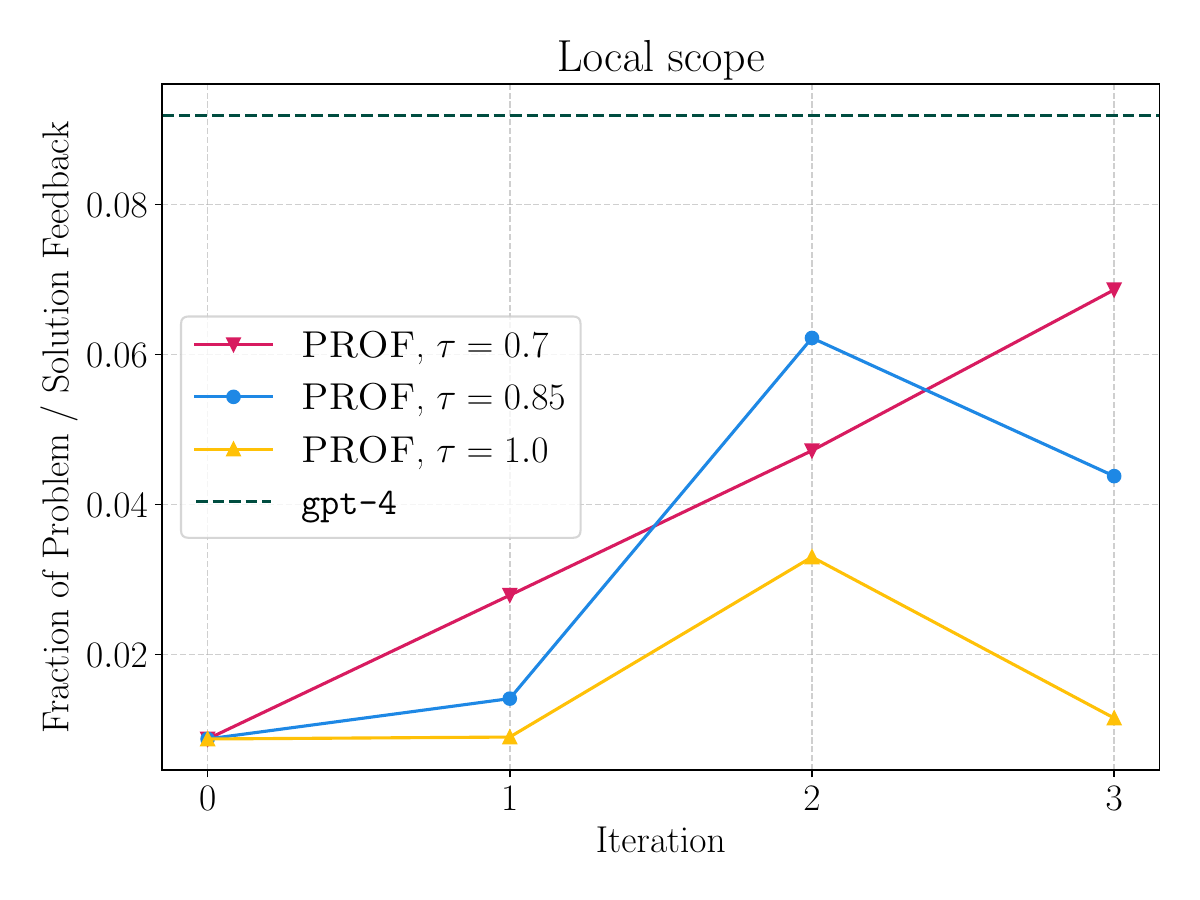}
\caption{Fraction of elements associated with local scope.}
\label{fig:ps_evolution_scope}
\end{subfigure}
\hfill
\begin{subfigure}[b]{0.33\textwidth}
\centering
\includegraphics[width=\textwidth]{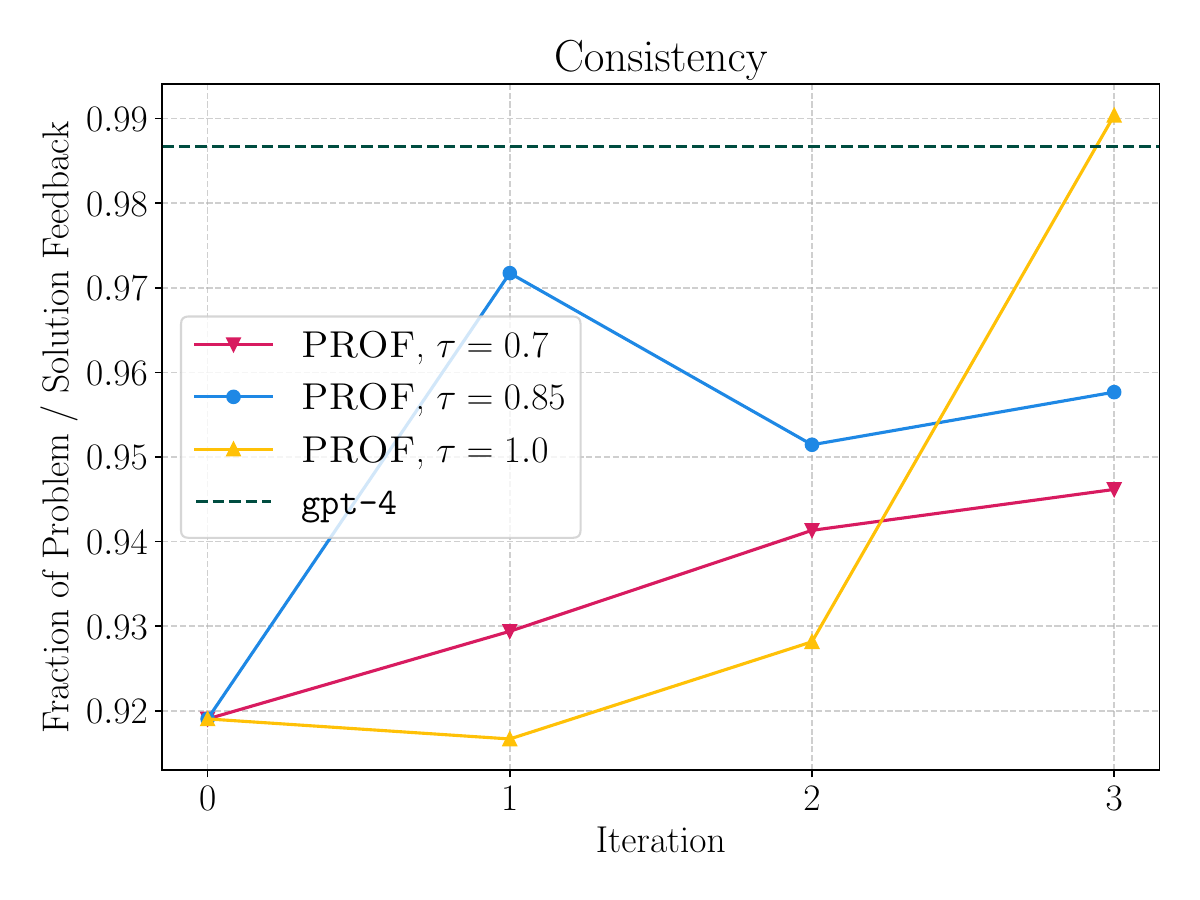}
\caption{Fraction of elements that are logically consistent.}
\label{fig:ps_evolution_consistency}
\end{subfigure}
\caption{Progression of problem/solution elements with refinement iterations.}
\label{fig:ps_evolution}
\end{figure}

% Our objective was to examine how the proportion of locally scoped feedback instances in the problem/solution segments changes with the number of refinement iterations to maintain an appropriate balance. 
Figure \ref{fig:ps_evolution_scope} illustrates that the fraction of locally-scoped feedback instances generally increases as the number of iterations grows, with the most significant increase observed for $\tau = 0.7$. This is desirable as initially the generated feedback has very few instances of locally-scoped elements and \model rectifies this by increasing it appropriately.  When the student simulator operates at a low temperature and makes minimal edits, the optimization algorithm encourages the feedback generator to prioritize generating feedback that focuses on local scope. This is because feedback associated with local scope is more likely to be addressed through localized changes, which align with the minimal edits made by the student simulator at low temperatures. 
% Conversely, when the student simulators are more liberal in refining the content, we observe a modest increase in the proportion of locally scoped feedback.

% Jiaye: I have a recommendation proofreading the writing. You say (3+3+1) * 8 = 56 revisions when you actually mean 56 revised essays; cause in the table above revision kind of mean an individual revision within one single essay, so this is a potential point of confusion.

\noindent \textbf{RQ3: Is the consistency of Problem / Solution segments improved with the number of refinement iterations?}
The \textbf{consistency} of a problem/solution is determined by two aspects: \textit{intrinsic correctness} and \textit{consistency with respect to the content}. Intrinsic correctness refers to the validity and absence of any logical fallacies in the feedback segment. 
% For example, if a solution segment includes reasoning, it must be valid and logically correct. 
Consistency with respect to the initial content refers to whether the identified problem is indeed an issue in the original content and whether the solution maintains the original stance of the essay without altering it. 
% This aspect ensures that the feedback remains consistent with the overall context and intent of the original writing.

Based on Figure \ref{fig:ps_evolution_consistency}, our training approach demonstrates an improvement in the consistency of the problem/solution segments as the number of refinement iterations increases. Notably, the consistency of \model, $\tau=1.0$ shows the highest performance after 3 iterations. We attribute this observation to the complexity of the training environment, which influences the consistency of the generated feedback. In the case of the student simulator executed with a temperature of 1.0, which yields lower implementation performance and creates a more challenging environment for the feedback generator, the training algorithm guides the feedback generator to produce feedback with better consistency to achieve optimal implementation performance. 
% Conversely, when the student simulator operates at a temperature of 0.7 and provides the best implementation performance during training, the training algorithm yields less improvement in the consistency of the generated feedback.

% \textbf{Takeaway:} The results presented in this section illustrate how the trained feedback generator successfully adapts to different student personas. When it comes to personas that require surface level modifications, our optimization algorithm directs the feedback generator to produce a higher quantity of feedback associated with local scope. For personas that struggle with implementation performance, the feedback generator becomes more cautious and delivers feedback that is logically consistent. Similarly, as the implementation performance decreases, our optimization process guides the feedback generator to generate fewer \textbf{problem} segments. Consequently, our algorithm proves effective in generating feedback and optimizing implementation performance for specific personas.

\section{Related Works}
% \lu{first discuss feedback generation, then preference optimization}
% \vspace{-2mm}
\subsection{Automatic Feedback Generation Systems}

NLP systems have been developed to automatically provide formative feedback to improve students' writing~\cite{liu2016automated,zhang2019erevise,klebanov2020automated}. One significant challenge faced by these approaches is the creation of high-quality feedback datasets, which requires considerable time and effort. In contrast, our approach starts with peer-annotated reviews that may not initially be of high quality. However, through an iterative preference learning process, we steer our feedback generator towards producing better quality responses. Considering that many previous works have focused on collecting peer review datasets~\cite{kang-etal-2018-dataset,lin2023moprd,dycke-etal-2023-nlpeer}, our approach can leverage these datasets for better initialization.
% \lu{not clear why this sentence is relevant, you mean these datasets can be used for initializing our generator? sinc our model doesn't require much feedback signal for training} 

While Language Models (LMs) can bypass the need for high-quality annotated feedback through few-shot prompting, they are computationally intensive and expensive~\cite{han2023fabric,chamoun2024automated,d2024marg}. In contrast, our feedback generator offers a cost-effective solution using smaller LMs with fewer parameters and without relying on high-quality supervision.

\subsection{Iterative Preference Optimization}

In recent times, a novel paradigm has emerged, which entails the iterative application of offline reinforcement learning techniques~\cite{NEURIPS2023_a85b405e}. In this approach, model generates preference relations per iteration, which are used to construct potentially more informative relations than those observed so far. This updates the model parameters, resulting in a better aligned model. Examples of such approaches include Iterative DPO~\cite{xu2023some,xiong2023gibbs,gulcehre2023reinforced}, Self-Rewarding LMs~\cite{yuan2024self}, and SPIN~\cite{chen2024self}. 
Previous approaches build preference relations by intrinsically evaluating the quality of the generation. However, exploring the utility of generation to establish preference relations has not been investigated before. To our knowledge, we introduce a method that constructs preference relations through extrinsic assessments, quantifying the utility of generated samples.

% Additionally, intrinsically assessing the effectiveness of generated feedback is challenging in our case due to the lack of a consensus on the essential attributes for improved effectiveness, thereby rendering the above approaches not directly applicable. Therefore, to the best of our knowledge, we are proposing the method of constructing preference relations through external assessment \lu{not sure is the right term, you mean evaluation based on the ultimate outcomes of importance} for the first time.
% \vspace{-2mm}
\section{Conclusion}
% \vspace{-2mm}
% \lu{conclusion too long}
In this paper, we present an optimization technique called \model that focuses on maximizing students writing revision performance through LM simulation. We conducted extensive analysis to showcase the alignment between our student simulators and actual revisions, while also demonstrating the flexibility of adjusting temperature to elicit diverse behaviors from the models. 
% This versatility can be harnessed to customize feedback generators and optimize implementation performance for specific personas, as well as create testing environments to evaluate the performance of feedback generators across a range of diverse personas. 
Through experiments, we show that our trained models not only exhibit better effectiveness but are more pedagogically aligned. 
% Through experiments, we not only observe improved effectiveness in our trained models, but also a stronger alignment with pedagogical principles.\lu{these two sentences are repetitive} 
% A more detailed analysis reveals that our method exhibits characteristics consistent with previous research aimed at enhancing feedback effectiveness.

\section*{Acknowledgements}
This work is supported in part through National Science Foundation under grant 2302564. We are grateful for the resources and services provided by Advanced Research Computing (ARC), a division of Information and Technology Services (ITS) at
the University of Michigan, Ann Arbor. Finally, we would like to extend our gratitude to Prof. Mitchell Dudley and Zhen Qian for their extensive contributions in designing the assignment, collecting student data and guiding us in comprehending the data.

\section*{Limitations}
We acknowledge the following limitations with our work:
\begin{itemize}
    \item As we consider a narrow domain for our experimental analyses, we concede that some of the conclusions and findings may not be directly extensible to other domains. 
    \item In our experimental setup, we employed LLMs to automatically assess the effectiveness of the generated feedback. As our analysis indicated a gap between some aspects of the simulated revision and the actual revision, we advise practitioners to approach the interpretation of the results with caution.
\end{itemize}

\section*{Ethical Statement}

As we rely on actual student data to train the LMs on simulating student implementation process, we affirm that the Institutional Review Board (IRB) approval was obtained prior to collecting and using student data, ensuring compliance with ethical standards for human subject research. Moreover, while our research simulates student implementation process, it should not be considered as a perfect representation of the actual student behavior. We strongly advise against using these simulations as a substitute for real student implementation procedures in educational settings. Instead, they should be viewed as a complementary tool to enhance understanding of student learning processes and to improve feedback mechanisms. 

% \lu{we need limitations and ethics? since we use llm to simulate human}

\bibliography{custom}

\appendix

\section{Dataset Prompts}
In this section, we attach all the prompts associated with the dataset described in \S \ref{sec:dataset_description}.

\subsection{Assignment Prompt}
\label{sec:assignment_prompt}
The prompt in Table \ref{tab:assignment_prompt} was shown to the students to get the initial writing:\\

\begin{table*}
    \caption{Assignment prompt shown to the students to get the initial writing.}
    \label{tab:assignment_prompt}
    \centering
    \begin{tabular}{p{0.95\linewidth}}
    \toprule
    \textbf{\textsc{Situation}} \\
    In San Francisco, the fear of losing jobs to automation after an increase to the minimum wage has motivated two similar policy proposals aimed at discouraging or banning automation. As a student of economics and someone who will soon enter the job market, you find this issue interesting and relevant. [For simplicity, assume these policies are only occurring in or are only proposed for the San Francisco area. Also assume the ban is for automation in general, not just delivery robots.] \\
    \textbf{\textsc{Prompt}} \\
    Write a letter to the San Francisco Board of Supervisors regarding the two policy proposals under consideration: a tax on automation (Ms. Kim’s proposal) or a ban on automation (Mr. Yee’s proposal). Your letter should: Briefly explain the economic impact of a minimum wage increase (moving from nonbinding to binding) in the labor market, and its extended effect on the automation market as well as the market for a good which may be produced using labor, automation or some combination of the two; Identify one of the two proposed policies and construct an argument, based on the economics you’ve learned in class, for why you oppose the policy. You, personally, may oppose both policy proposals, but your paper should focus on only one policy, given the word count limit. Your argument of opposition should not be based on your support for the other proposed policy. While your letter is a normative economic assessment, the majority of the letter should consist of positive economic analysis. [While you may have strong opinions on this subject based on moral or ethical reasoning, the purpose of this assignment is to see your ability to use the economic tools you’ve learned to analyze the situation.] Explain the economic impact of this policy proposal on these same markets, highlighting the economic reasoning for opposing the policy; Start your analysis assuming the minimum wage increase already occurred. Be persuasive. \\
    \textbf{\textsc{Items to keep in mind:}} \\
    The Board of Supervisors likely has some knowledge of economics. Your explanations may assume prior general knowledge consistent with our coverage of Supply, Demand, Consumer Surplus, Producer Surplus, and Efficiency. The supervisors understand the definitions of these terms, but not necessarily how they interact specifically with this problem. For example, the Board of Supervisors do not immediately understand how a price change in the labor market affects related markets. The provided article must be cited. External references are not required but also must be cited if used. Either APA format or MLA format is acceptable. Since you are writing to the Board of Supervisors, you should take care to carefully edit and proofread your letter. Your letter should be between 400-500 words (this includes your first draft) and should follow the conventions of a professional letter, including a To and From section as well as a professional and courteous tone. Please sign your letter “A Concerned Citizen” since anonymity is needed for peer review. Please include a word count at the end of your assignment. [The word count may exclude the “To” and “From” lines, as well as the parenthetical citations within your paper. Note quotations within your paper still count. Don't forget that both the minimum and maximum limits to the word count are important on all drafts.] \\
    \bottomrule
    \end{tabular}
\end{table*}

\subsection{Peer Review Prompt}
\label{sec:peer_review_prompt}
The prompt in Table \ref{tab:peer_review_prompt} was shown to the students while reviewing the initial writing:\\

\begin{table*}
    \centering
    \caption{Prompt shown to the students to review an initial writing.}
    \label{tab:peer_review_prompt}
    \begin{tabular}{p{0.95\linewidth}}
    \toprule
    In providing feedback, your task is to help your peers identify areas that need the most attention. To guide you through the process of effectively providing feedback according to the essay rubric, you will be given a series of prompts that correspond to the essay rubric criteria.\\
    \textbf{\textsc{Peer review guidelines}}
    \begin{itemize}
        \item Highlight the pieces of texts that let you directly address the feedback prompts in your online responses.
        \item In your online responses, focus on larger issues (higher order concerns) of content and argument rather than lower order concerns like grammar and spelling.
        \item Be very specific in your responses, referring to your peer’s actual language, mentioning terms and concepts that are either present or missing, and following the directions in the feedback prompts.
        \item Use respectful language whether you are suggesting improvements to or praising your peer.
    \end{itemize}
    \textbf{\textsc{Peer review feedback prompts:}}\\
    
    \textbf{Understanding}:
    \begin{itemize}
        \item Based on your class discussion and course readings, identify any important concepts that are missing. Identify any unnecessary concepts in use.
        \item How can the author connect concepts in a more useful manner? For example, using your knowledge from class, how could the author improve their explanation of interactions between the various markets affected by these policies?
    \end{itemize}
    \textbf{Critical Thinking}
    \begin{itemize}
        \item Based on your class discussion and course readings, how could the author improve their analysis of the minimum wage increase and the automation policy (ban or tax, depending)?
        \item How well does the author apply economic principles to justify his/her position? Suggest one (or two) additional ways the author could apply economics to their argument in order to make this letter more persuasive.
        \item Are all outside sources properly cited?
    \end{itemize}
    \textbf{Response Alignment with Audience}
    \begin{itemize}
        \item The letter should be understandable to a person with a basic but not sophisticated understanding of economic principles. In this context, which parts were difficult to understand? Which parts were easy to understand?
    \end{itemize}\\
    \bottomrule
    \end{tabular}
\end{table*}

\subsection{Prompt for Grading Essays}
\label{sec:grade_essay_prompt}

 The prompt in Table \ref{tab:grade_essay_prompt} includes the rubric to grade an essay. On a high level, it elucidates the criteria for assigning a particular point for each of the question / instruction in the peer review prompt from Appendix \ref{sec:peer_review_prompt}. The detailed rubric is given as follows:

\begin{table*}
    \small
    \centering
    \caption{This prompt provides the rubric for grading the essay.}
    \label{tab:grade_essay_prompt}
    \begin{tabular}{p{0.95\linewidth}}
    \toprule
    \begin{itemize}  
    \item \textbf{\textsc{Understanding}}  
    \begin{itemize}  
        \item \textbf{Concepts \& Accuracy}:   
        \begin{itemize}  
             \item Missing elements: Missing several central economic terms and/or correct definitions. (1, 2, 3 Points)  
             \item Meets expectations: Almost all economic concepts central to the policies and markets are identified and/or correctly defined. (4 Points)  
             \item Exceeds Expectations: All economic concepts central to the policies and markets are identified and correctly defined in a way that exceeds expectations for the course. (5 Points)  
         \end{itemize}  
        \item \textbf{Linking Concepts}:  
        \begin{itemize}  
            \item Missing elements: Several connections between relevant concepts and markets are missing or incorrect. (1, 2, 3 Points)  
            \item Meets expectations: Mostly correct connections between relevant concepts and markets in a way that demonstrates an understanding consistent for the course. (4 Points)  
            \item Exceeds Expectations: Building upon their definitions, the writer correctly connects the relevant concepts and markets to one another demonstrating an understanding that is sophisticated for the course. (5 Points)  
        \end{itemize}  
        \item \textbf{Conciseness}:  
        \begin{itemize}  
            \item Missing elements: Over word count (>510 words), with a large number of sentences and/or words that are not directly related to the prompt. (1, 2, 3 Points)  
            \item Meets expectations: Within the word count (up to 510 words), with places in which descriptions are wordy, suggesting a lack of understanding.  
            \item Exceeds Expectations: Able to answer question succinctly in less than 500 words.  
        \end{itemize}  
    \end{itemize}  
    \item \textbf{\textsc{Critical Thinking}}  
    \begin{itemize}  
        \item \textbf{Interpreting Sources}:  
        \begin{itemize}  
            \item Missing elements: Interpretation suggests a limited understanding of the economics within the source(s); Writing suggests source(s) may not have been read; No citations. (1, 2, 3 Points)  
            \item Meets expectations: Interprets the economics within the article consistent with expectations for the course while predominantly quoting the source(s); An attempt to cite source(s) is made. (4 Points)  
            \item Exceeds Expectations: Accurately interprets and articulates the economics within the source(s) in a sophisticated manner while predominantly summarizing source(s); citations properly formatted. (5 Points)  
        \end{itemize}  
        \item \textbf{Analysis of Case Study}:  
        \begin{itemize}  
            \item Missing elements: Opposition to a policy is missing; Or the argument does not acknowledge the economic impact of the policy. (1, 2, 3 Points)  
            \item Meets expectations: Argument effectively debunks the proposed solution in a manner consistent with the level of the class. Full and thorough articulation of each market interaction is not present. (4 Points)  
            \item Exceeds Expectations: Insightful articulation of the issues facing one of the proposed solutions. All market interactions are explored, coming to a conclusion indicating that the proposed solution is not economical. Accurately interweaves each economic concept present in the proposal into their articulation of the downsides. (5 Points)  
        \end{itemize}  
    \end{itemize}  
    \item \textbf{\textsc{Response Alignment With Audience}}  
    \begin{itemize}  
        \item Missing elements: Explanations do not align with the expected audience; Recommendations are inconsistent with the target audience, but carry reasonable economic analysis. Ex: Recommending government action when the audience is producers. (1, 2, 3 Points)  
        \item Meets expectations: Explanations generally align with audience needs but tend to be too advanced or too simple for the specific audience; Format mostly correct. (4 Points)  
        \item Exceeds Expectations: Assumes format described in prompt and explains concepts in a way that consistently meets audience needs. (5 Points)  
    \end{itemize}  
    \end{itemize}\\
    \bottomrule
    \end{tabular}
\end{table*}

\subsection{Prompt for Combining Feedback}
\label{sec:combining_feedback_prompt}
This prompt used for combining the $3$ peer reviews into a single review that captures all the comments from the individual feedback is shown below:

\begin{quote}
    Your task is to skillfully merge feedback from multiple reviewers into one unified and coherent narrative. Maintain the original language style and tone of each review to accurately represent their feedback. In the following examples, you will be provided with input from several reviewers, and your job will be to craft a single piece of feedback that honors the format and intent of the individual contributions, without explicitly mentioning reviewer numbers.
\end{quote}

\begin{table*}
    \centering
    \scriptsize
    \caption{Example of 3 peer reviews. The results for combining this into a single peer review is shown in Table \ref{tab:combined_peer_review}.}
    \label{tab:3_peer_review}
    \begin{tabular}{p{0.95\linewidth}}
        \toprule
        \textsc{\textbf{Review 1}}:\\
        {\bf Understanding 1: Based on your class discussion and course readings, identify any important concepts that are missing. Identify any unnecessary concepts in use.:} The author did not include anything about non-binding and what it has to do with minimum wage. Other than that, the author did not miss anything and did not include anything extra. \\
        {\bf Understanding 2: How can the author connect concepts in a more useful manner? For example, using your knowledge from class, how could the author improve their explanation of interactions between the various markets affected by these policies?:} The author can discuss substitutes relevant to automation to better connect their topic. Additionally, the author can talk more about the labor market which will help connect the job loss from automation. \\
        {\bf Critical Thinking 1: Based on your class discussion and course readings, how could the author improve their analysis of the minimum wage increase and the automation policy (ban or tax, depending)?:} I liked how the author talked about the graph in explaning these concepts, but I believe they could bring in real-world examples to help strengthen their argument and give the reader better understanding. Additionally, the author can talk a little bit more about the long term effects of automation.\\
        {\bf Critical Thinking 2: How well does the author apply economic principles to justify his/her position? Suggest one (or two) additional ways the author could apply economics to their argument in order to make this letter more persuasive.:} Like I said before, the author can incorporate more effects of automation which would help influence the reader to take their side rather than oppose it. \\
        {\bf Critical Thinking 3: Are all outside sources properly cited?:} No, it is not cited correctly. There are punctuation and capitalization errors.
        {\bf Response Alignment with Audience:} The letter should be understandable to a person with a basic but not sophisticated understanding of economic principles. In this context, which parts were difficult to understand? Which parts were easy to understand?: The author should explain the concept of deadweight loss better. However, everything else in the paper is explained in the proper manner and is easy to understand.\\
        \midrule
        \textsc{\textbf{Review 2}}:\\
        {\bf Understanding 1: Based on your class discussion and course readings, identify any important concepts that are missing. Identify any unnecessary concepts in use.:} There are no missing terms or important details. \\
        {\bf Understanding 2: How can the author connect concepts in a more useful manner? For example, using your knowledge from class, how could the author improve their explanation of interactions between the various markets affected by these policies?:} I think that some explanations could be more clearer and easier to understand. For example, when you say that a binding price control will ensure that the market will not reach equilibrium.  You could first mention what a binding price control is and also explain how that applies to minimum wage. Another place where you could connect ideas better is when you mention that "This surplus causes deadweight loss and inefficiency", even though you mentioned that the market is below the equilibrium before. I feel like the connection between that and the inefficiency is not clear. Instead, you could explain that the market is below equilibrium, which causes inefficiency immediately after or before you say that the surplus causes deadweight loss and inefficiency. \\
        {\bf Critical Thinking 1: Based on your class discussion and course readings, how could the author improve their analysis of the minimum wage increase and the automation policy (ban or tax, depending)?:} One thing that could help you improve your analysis on automation is the effects of society. In your paper, you only mentioned the effects of the labor market and the employers, you never really mention the effects of these changes on society. Adding this could help improve your analysis of automation and allow you to look at more negative factors.\\
        {\bf Critical Thinking 2: How well does the author apply economic principles to justify his/her position? Suggest one (or two) additional ways the author could apply economics to their argument in order to make this letter more persuasive.:} One thing that you could do to improve your argument would be going more in-depth into the markets. For example, when you can mention the effect of the goods when you want to automate or not. For example, when you have automation, you will produce more goods and will therefore drop the price. Compared to a scenario which does not automate, they make less goods, which will have an increased price compared to the one with automation. Citizens will more likely want to purchase the cheaper option, which opposes the ban on automation.\\
        \midrule
        \textsc{\textbf{Review 3:}} \\
        {\bf Understanding 1: Based on your class discussion and course readings, identify any important concepts that are missing. Identify any unnecessary concepts in use.:} You talked about all the topics and I think you layed out your essay nicely. However, I would expand upon the influence of the policies/concepts more. I don’t think it’s necessary to introduce producer surplus in this essay, or at least not in the way it’s currently pulled in. \\
        {\bf Understanding 2: How can the author connect concepts in a more useful manner? For example, using your knowledge from class, how could the author improve their explanation of interactions between the various markets affected by these policies?:} I think adding to your second and fourth paragraphs by diving into the specific effects of each policy would strengthen your explanations about the connections between each market. The fourth paragraph is kind of confusing and I feel like you may be extrapolating past the immediate effects. \\
        {\bf Critical Thinking 1: Based on your class discussion and course readings, how could the author improve their analysis of the minimum wage increase and the automation policy (ban or tax, depending)?:} I couldn’t tell which demand curve shift you were describing at the start of your 4th paragraph. I think it shows your shift to talking about the automation market, but it’s not super clear. I think it would add to your argument to talk more about the unemployment that would be caused by banning automation for all the workers involved in that market and those in the companies forced to close due to high labor prices. \\
        {\bf Critical Thinking 2: How well does the author apply economic principles to justify his/her position? Suggest one (or two) additional ways the author could apply economics to their argument in order to make this letter more persuasive.:} I think pointing out that labor and automation are substitutes was really smart and helped illustrate the point you’re making. A few effects are misattributed. For example, the deadweight loss is a result of the price floor, not the surplus. Also, instead of quoting Dr. Dudley’s slides, I would rephrase those definitions in your own words to demonstrate understanding. \\
        {\bf Critical Thinking 3: Are all outside sources properly cited?:} They properly cited the article provided and another outside source with in-text citations and a correct works cited page. But, you don’t need to cite Dr. Dudley’s slides in this paper.\\
        {\bf Response Alignment with Audience:} The letter should be understandable to a person with a basic but not sophisticated understanding of economic principles. In this context, which parts were difficult to understand? Which parts were easy to understand?: The assignment does suggest you sign off with a concerned citizen, but your sign-off works as well, but I’d change “economic” to ‘economics’. You have some grammatical errors scattered throughout the essay, fixing those may make the paper a little easier to read and more professional given the audience. But overall the order of your essay and the stance you take is well-founded. Good job! \\
        \bottomrule
    \end{tabular}
\end{table*}

\begin{table*}
    \centering
    \small
    \caption{The result of combining the 3 peer reviews shown in Table \ref{tab:3_peer_review} using the prompt shown in \ref{sec:combining_feedback_prompt}.}
    \label{tab:combined_peer_review}
    \begin{tabular}{p{0.95\linewidth}}
    \toprule
         \textbf{\textsc{Output from} \texttt{gpt-3.5}}\\
{\bf Understanding 1: Based on your class discussion and course readings, identify any important concepts that are missing. Identify any unnecessary concepts in use.:} The author did not include anything about non-binding and what it has to do with minimum wage. Additionally, while you talked about all the topics and laid out your essay nicely, expanding on the influence of the policies/concepts more would be beneficial. There are no missing terms or important details overall, but it’s not necessary to introduce producer surplus in this essay, or at least not in the way it’s currently pulled in.\\
{\bf Understanding 2: How can the author connect concepts in a more useful manner? For example, using your knowledge from class, how could the author improve their explanation of interactions between the various markets affected by these policies?:} The author can discuss substitutes relevant to automation to better connect their topic. Additionally, the author can talk more about the labor market to help connect the job loss from automation. Some explanations could be clearer and easier to understand. For example, when you say that a binding price control will ensure that the market will not reach equilibrium, you could first mention what a binding price control is and explain how that applies to minimum wage. Strengthening the second and fourth paragraphs by diving into the specific effects of each policy would also help. Clarifying that the market being below equilibrium causes inefficiency would make the connection between surplus, deadweight loss, and inefficiency clearer. \\
{\bf Critical Thinking 1: Based on your class discussion and course readings, how could the author improve their analysis of the minimum wage increase and the automation policy (ban or tax, depending)?:} I liked how the author talked about the graph in explaining these concepts, but incorporating real-world examples would help strengthen the argument. One thing that could help improve your analysis on automation is discussing the effects on society, not just the labor market and employers. Additionally, discussing more about the long-term effects of automation and the unemployment caused by banning automation would be beneficial. Clarifying which demand curve shift you are describing in the fourth paragraph would also add clarity.\\
{\bf Critical Thinking 2: How well does the author apply economic principles to justify his/her position? Suggest one (or two) additional ways the author could apply economics to their argument in order to make this letter more persuasive.:} The author can incorporate more effects of automation to help influence the reader to take their side rather than oppose it. Going more in-depth into the markets, such as mentioning the effects on goods prices with and without automation, would strengthen the argument. Pointing out that labor and automation are substitutes was effective, but correctly attributing deadweight loss to the price floor rather than the surplus, and rephrasing definitions from class materials in your own words, would improve the application of economic principles.\\
{\bf Critical Thinking 3: Are all outside sources properly cited?:} No, it is not cited correctly. There are punctuation and capitalization errors. You don’t need to cite Dr. Dudley’s slides in this paper. \\
{\bf Response Alignment with Audience: The letter should be understandable to a person with a basic but not sophisticated understanding of economic principles. In this context, which parts were difficult to understand? Which parts were easy to understand?:} The author should explain the concept of deadweight loss better. However, everything else in the paper is explained in a proper manner and is easy to understand. There are some typos in your paper, such as misspelling "substitute". While the order of your essay and your stance are well-founded, fixing grammatical errors and using professional language will make the paper easier to read and more aligned with the audience. The assignment suggests signing off as "a concerned citizen," but your sign-off works as well. Good job!\\
    \bottomrule
    \end{tabular}
\end{table*}

Using the above prompt on 3 peer review shown in Table \ref{tab:3_peer_review} results in the combined feedback shown in Table \ref{tab:combined_peer_review}.

\section{Implementation Details: Hyperparameter Settings}
\label{sec:hyperparameter_settings}
For training the student simulator and initializing the feedback generator using \texttt{llama-8b-instruct}, we use LoRA~\cite{hu2021lora} with a batch size of $8$ and cosine annealing with warm restarts~\cite{loshchilov2016sgdr} for linear rate scheduling. The model undergoes 20 epochs of training, with an initial learning rate of $10^{-4}$.

Our feedback generator is iteratively refined for $3$ iterations. At each stage of sampling preference relations, we sample $K=5$ different feedback for each datapoint. For the DPO optimization, we set $\beta = 0.1$ and train the model under this framework for $5$ epoches with an initial learning rate of $10^{-5}$ and a batch size of $8$.

For each of the above described training processes, we use \texttt{LLaMA-Factory}~\cite{zheng2024llamafactory}.

\section{Student Simulator: Additional Analysis}

\subsection{Temperature and Word-level Modifications}
\begin{figure}[h]
     \centering
     \includegraphics[scale=0.3]{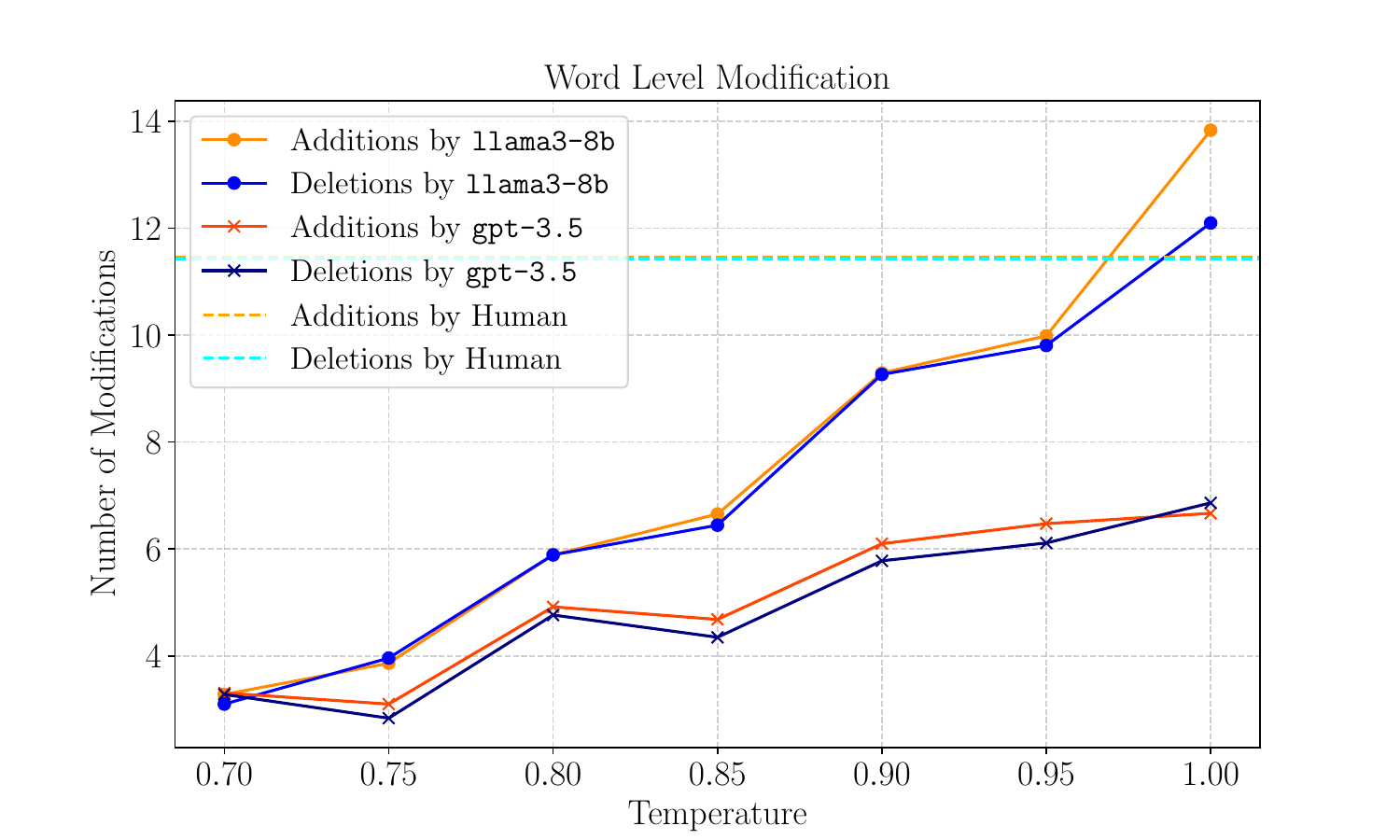}
     \caption{Temperature and Word-level modifications}
     \label{fig:num_modifications_word_level}
 \end{figure}

The variation of the number of modified by the student simulators with temperature is shown in Figure \ref{fig:num_modifications_word_level}.

\begin{table}[h]
    \setlength\tabcolsep{12pt}
    \scriptsize
    \centering
    \begin{tabular}{c|ccc}
        \midrule
        \multirow{3}{*}{\textsc{Approach}} & \multicolumn{3}{c}{\textsc{Student Simulators}} \\
        & \multicolumn{3}{c}{\texttt{llama3-8b}} \\
        & 0.7 & 0.85 & 1.0 \\
        \midrule
        \texttt{gpt-3.5} & 75.3 & 75.7 & 72.4 \\
        \texttt{gpt-4} & \best{78.3} & 76.4 & 75.2 \\
        \texttt{sft-from-human} & 73.7 & 74.5 & 73.4 \\
        \midrule
        \multicolumn{4}{c}{\model, $\tau = 0.7$} \\
        \midrule
        \textit{Iteration 1} & 76.9 & 	76.4 & 	\best{75.5}	\\
        \textit{Iteration 2} & 76.3 &	76.8 &	75.2	\\
        \textit{Iteration 3} & 76.4 & 	\secondbest{77.1} & 	74.7 \\
        \midrule
        \multicolumn{4}{c}{\model, $\tau = 0.85$} \\
        \midrule
        \textit{Iteration 1} & 76.6 & 	76.5 & 	72.0\\
        \textit{Iteration 2} & 76.5 &	76.2 & 	73.6\\
        \textit{Iteration 3} & \secondbest{77.5} & 	\best{77.8} & 	72.0 \\
        \midrule
        \multicolumn{4}{c}{\model, $\tau = 1.0$} \\
        \midrule
        \textit{Iteration 1} & 73.5 &	76.4 &	75.1 \\
        \textit{Iteration 2} & 76.3 &	76.3 &	68.4\\
        \textit{Iteration 3} & 75.7 &	75.1 &	\secondbest{75.3}\\
        \bottomrule
        
    \end{tabular}
    \caption{
    Extrinsic evaluation of the generated feedback in terms of simulated revision performance. \best{Green} and \secondbest{Blue} represents best and second-best performance respectively. Each experiment was repeated for $5$ different seeds to mitigate the impact of randomness.}
    \label{tab:extrinsic_evaluation_llama}
\end{table}

\subsection{Qualitative Analysis of Revisions}
\label{sec:ss_analysis_revision}

Upon analyzing the individual data points and corresponding revisions from both simulators, we have observed that while \texttt{llama3-8b} generates a higher number of revisions, these changes do not necessarily enhance the quality of the initial writing. Specifically, we have noticed that it often deletes crucial statements that are essential for the overall argumentation of the essay, while also introducing statements that may negatively impact clarity and persuasiveness. We present a qualitative example in this section to illustrate undesirable and desirable feedback implementation from the \texttt{llama3-8b} and \texttt{gpt-3.5} based student simulator respectively.

\begin{figure*}[h]
    \centering
    \includegraphics[scale=0.65]{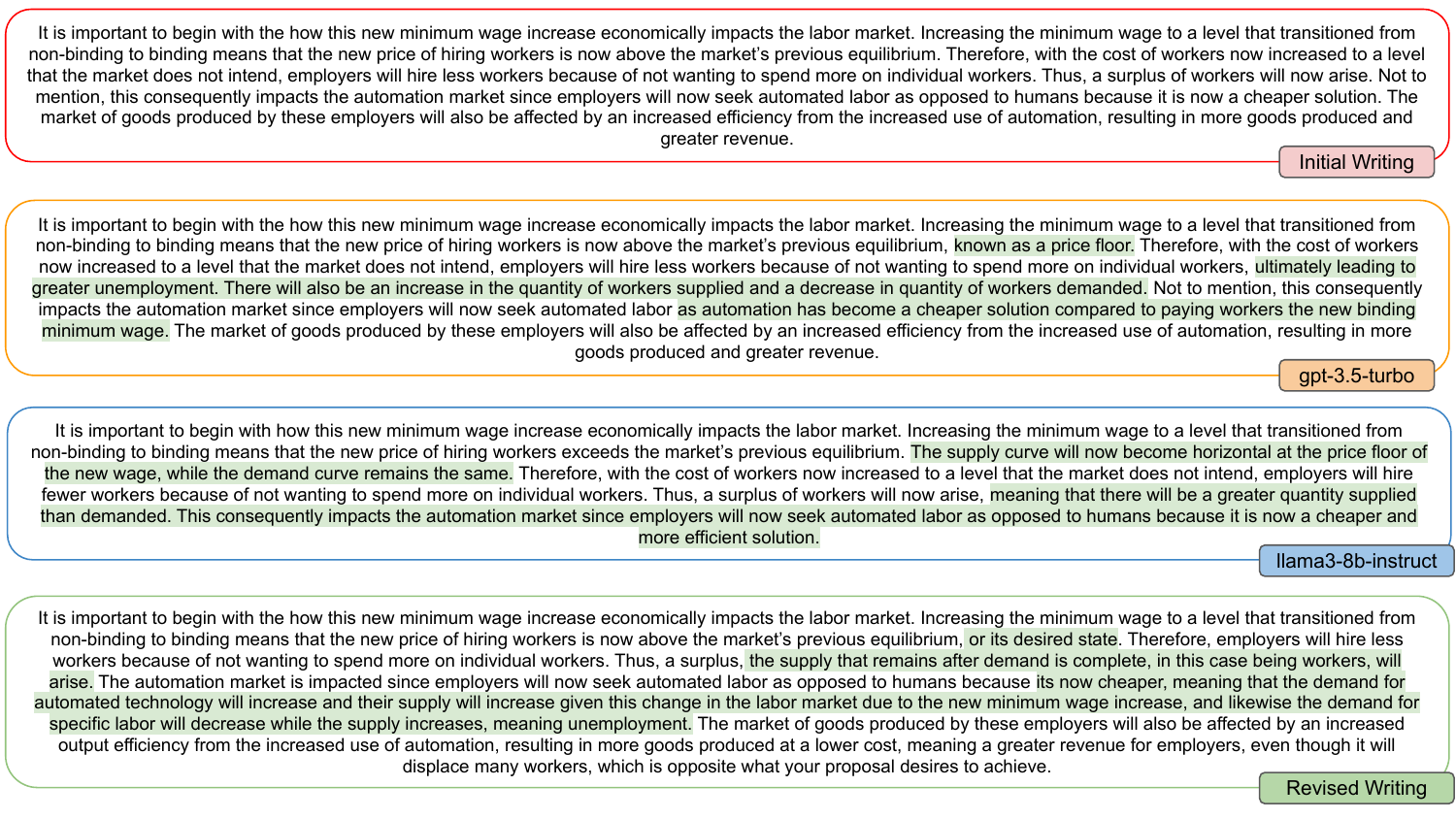}
    \caption{This figure illustrates the modifications made to a paragraph in the initial writing by both student simulators and a real student. It specifically highlights the ineffective changes made by the \texttt{llama-8b-instruct} simulator, in contrast to the improvements made by the \texttt{gpt-3.5} simulator, which ultimately degrade the overall quality of the essay. For further insights into the subpar performance of the \texttt{llama-8b-instruct} based student simulator, please refer to \S \ref{sec:ss_analysis_revision}.}
    \label{fig:ss_analysis_revision}
\end{figure*}

As shown in the Figure \ref{fig:ss_analysis_revision}, the revised version from \texttt{gpt-3.5} based student simulator retains the more detailed and comprehensive explanation of how the minimum wage increase affects the labor market (\textit{"Increasing the minimum wage ... known as a price floor."}). It explicitly mentions the concept of a price floor and explains how it leads to higher unemployment which in turn results in an increase to the quantity of the workers supplied at a lower quantity of the workers demanded (\textit{"Therefore, with the cost of workers
now increased to a level ... ultimately leading to
greater unemployment. ... in quantity of workers demanded."}). Overall, the improved version provides a clearer and more nuanced understanding of the economic implications of the minimum wage increase on the labor market and this improves the persuasiveness of the argument. 

On the other hand, the changes implemented by \texttt{llama3-8b} results in the deletion of an important line that explains why employers would prefer automation market (\textit{"The market of goods produced by these employers ... more goods produced and greater revenue."}). While it makes the paragraph more technical by explaining how the supply curve and the demand curve is impacted using the concept of \textit{price floor}(\textit{"The supply curve will ... demand curve remains the same."}), it is unclear and not immediately deducible from the previous statements. This negatively impacts the clarity and persuasiveness of the argument.

\section{Analyses of Feedback Generation: Additional Details}

\subsection{Extrinsic Evaluation using \texttt{llama3-8b}}
\label{sec:extrinsic_evaluation_llama}

In this section, we attach the effectiveness of the generated feedback using \texttt{llama-3b} in Table \ref{tab:extrinsic_evaluation_llama}. On a high-level, we notice that our approaches perform better than the few-shot approaches involving enterprise LLMs and \texttt{sft-from-human}. This indicates that for behavior traits associated with poor revision performance, our approaches are more effective.

\subsection{Example of Scoring a Feedback along Pedagogical Dimensions}
\label{sec:example_pedagogical}

In this section, we attach the output from \texttt{gpt-4} which contains the analysis of a feedback along pedgagogical aspects in Table \ref{tab:example_pedagogical_feedback}.

\begin{table*}
    \centering
    \scriptsize
    \caption{Example of a feedback evaluated along pedagogical aspects by \texttt{gpt-4}.}
    \label{tab:example_pedagogical_feedback}
    \begin{tabular}{p{0.95\linewidth}}
    \toprule
        { \sc \textbf{Feedback:}}\\ 
{\bf Understanding 1: Based on your class discussion and course readings, identify any important concepts that are missing. Identify any unnecessary concepts in use.:} As I was reading through this letter, I noticed that many fundamental concepts were missing. For example, the author forgot to mention that an increase in the minimum wage, set by a binding price floor, results in a surplus of labor, as quantity supplied is greater than quantity demanded. Also, if the price of labor goes up in tandem with an imposition of taxes on automation, then the supply of goods and services produced using labor and automation will decrease, causing equilibrium price and quantity to increase and decrease respectively. Additionally, the author failed to mention how consumer, producer, and total surplus would be affected by the tax, which is a crucial component of this prompt.\\
{\bf Understanding 2: How can the author connect concepts in a more useful manner? For example, using your knowledge from class, how could the author improve their explanation of interactions between the various markets affected by these policies?:} To reiterate, I felt that while the author does a good job at talking about the markets individually, they do not talk about how they are all interconnected. The author can explain that the change in minimum wage would not only cause a surplus in labor, but an increase in the demand for automation, because labor and automation are most possibly substitutes in the production of particular goods and services. When the ban is imposed on automation, the dynamics between these markets become much more complex. For example, in the market for goods and services, the supply of these goods and services now decreases, reducing equilibrium quantity while simultaneously increasing the equilibrium price. Therefore, not only would suppliers pay, in the form of added input costs, but so would consumers, who are now forced to pay a higher price to receive the same goods that were cheaper before the tax on automation.\\
{\bf Critical Thinking 1: Based on your class discussion and course readings, how could the author improve their analysis of the minimum wage increase and the automation policy (ban or tax, depending)?:} The author gave a very brief description of how an increase in the minimum wage affects markets, such as the labor market, by stating that it “triggers various economic repercussions.” beyond that, they did not, but should explain how it creates a surplus of labor in the labor market, as laborers are enticed by a higher wage, while employers are deterred, because they see it as a potential rise in input costs, which will significantly affect profits. They can also improve their analysis of the tax on automation by explaining how it may affect both producers and consumers of the goods and services produced by labor and automation. Hence, both producers and consumers of these goods and services may experience a decrease in producer and consumer surplus.\\
{\bf Critical Thinking 2: How well does the author apply economic principles to justify his/her position? Suggest one (or two) additional ways the author could apply economics to their argument in order to make this letter more persuasive.:} In my opinion, I feel that there was a lot more the author could have done to effectively justify his/her position. It appeared to me that the author’s primary argument against taxation was that producers will still continue to use automation regardless of a tax, which is not necessarily true. To advance their argument, the author could have spoken from an efficiency standpoint, explaining that total welfare is prevented from being maximized as a result of the tax on automation; this would occur, because of either a decrease in consumer or producer surplus or both. Also, they could have argued that automation acts as a lifeline for producers who need a cheaper and more efficient alternative to labor, and adding the tax on automation would force many of these producers to exit the market and stop producing.\\
{\bf Critical Thinking 3: Are all outside sources properly cited?:} From my understanding, the writer used the Saltsman article on the wage floor, but did not accurately cite it. They could resolve this by citing the article on a work cited page.\\
{\bf Response Alignment with Audience:} The letter should be understandable to a person with a basic but not sophisticated understanding of economic principles. In this context, which parts were difficult to understand? Which parts were easy to understand?: In all honesty, while the content discussed in the letter was pretty comprehensible, I found the general argument against the tax on automation to be confusing. For instance, the author stated that, “If we only put a tax on automation, automation will still replace the labor force.” However, I believe that this truly depends on the magnitude of the tax on automation; if producers can generate more profit using human labor than automation, then they will use labor, and not automation like the author implied.” On a more positive note, a part that I thought was very well explained was the potential solutions to the rising minimum wage, such as investing in education. \\
\midrule
\textbf{\textsc{Score from \texttt{gpt-4}}:}\\
\textbf{Respects Guided Question:} The feedback partially aligns with the guided questions by providing a detailed critique of missing economic concepts and their interconnections, as well as suggestions to strengthen the essay's arguments. However, it falls short in directly addressing some specific guided questions, like the precise connection between the minimum wage increase and automation policies. The feedback touches on these topics but doesn't fully explore their interrelation as prompted. (2 Points) \\
\textbf{Encourages Active Learning:} The feedback encourages the author to think more deeply about the economic implications and connections between concepts like labor market surplus and the effects of automation tax. Suggestions for linking these concepts and exploring their impact on both producers and consumers promote a more active engagement with the material. However, the feedback sometimes outlines specific deficiencies and corrections rather than prompting the author to identify these gaps independently. (3 Points) \\
\textbf{Deepens Metacognition}: The feedback effectively identifies key areas where the student's understanding and application of economic principles are lacking, such as the impact of minimum wage increases on labor surplus and the overall economic system. It calls attention to the need for a deeper analysis of how these factors interact, thus helping the student recognize the complexities of economic policies. However, it could further enhance metacognition by asking reflective questions that lead the student to evaluate their reasoning process. (4 Points) \\
\textbf{Motivates and Stimulates Student Curiosity}: The tone of the feedback is constructive and focuses on areas for improvement, which can motivate the student to delve deeper into economic theories. However, it lacks elements that explicitly encourage curiosity, such as posing intriguing questions or suggesting novel viewpoints that could expand the student’s perspective on the topic. (3 Points) \\
\textbf{Adapts to Essay Quality}: The feedback is well-tailored to the quality of the essay. It recognizes the student’s basic understanding and accordingly provides detailed explanations and corrections that aim to elevate the student's analysis from a simplistic view of economic impacts to a more nuanced understanding. The feedback is specific where needed but also pushes for a higher-level synthesis of economic interactions. (4 Points) \\
    \bottomrule
    \end{tabular}
\end{table*}

\section{Annotation Guidelines: Student Simulator Faithfulness}
\label{sec:ss_ann}
The detailed guidelines for conducting student simulator faithfulness analysis is provided in Table \ref{tab:ss_ann}.

\begin{table*}
    \centering
    \caption{Annotation Guidelines for conducting student simulator faithfulness analysis}
    \label{tab:ss_ann}
    \begin{tabular}{p{0.95\linewidth}}
        \toprule
         Your task is to assess the degree to which the student simulator incorporates the feedback generated. Please adhere to the outlined steps sequentially to mitigate potential errors and biases.
         \begin{enumerate}
             \item Annotate the feedback: Each essay is accompanied by three feedback responses from student peer evaluators, all pertaining to the \textbf{SAME} essay. These responses may contain repetitions either across the three feedback or within a single feedback. As you read through the feedback, mark each unique suggested revision for annotation.

             \begin{itemize}
                \item Highlight each new suggestion (i.e., recommended revision) you encounter using a distinct color. Ensure your chosen color is as different as possible from others. Immediately following the highlighted sentence, provide a succinct summary of the suggestion in parenthesis.
                
                % \sethlcolor{yellow}
                % \begin{enumerate}
                %     \item {\small Example: \hl{They did not include the ideas of consumer and producer surplus.}} (\textit{\small Suggestion1: discuss consumer/producer surplus})
                % \end{enumerate}
            
                \item If the suggestion (i.e., recommended revision) has already been proposed either within the same feedback or in a separate feedback out of the three peers’ feedback, shade it in gray. Do \textbf{NOT} consider this as new feedback. Immediately following the shaded sentence, paste your previous summary of the suggestion (or a combination of suggestions).
                % \sethlcolor{gray!30}
                % \begin{enumerate}
                %     \item {\small Example: \hl{The author can explain that the change in minimum wage would not only cause a surplus in labor, but an increase in the demand for automation, because labor and automation are most possibly substitutes in the production of particular goods and services.}} (\textit{\small Suggestion6: discuss effect of min wage on labor market + Suggestion2: discuss effect of min wage on supply/demand of automation market})
                % \end{enumerate}
            \end{itemize}

            \item Put together all distinct suggestions into a list. And document the number of recommended changes (i.e. suggestions) in the provided table.
            {\small Example: }
            \begin{itemize}
                \item {\small Suggestion1: discuss consumer/producer surplus}
                \item {\small Suggestion2: discuss equilibrium}
                \item {\small Suggestion3: discuss effect of min wage on supply/demand of automation market}
            \end{itemize}
            \item Evaluate the revised essay by the real student, simulated student (temp = 0.70), simulated student (temp = 0.85), simulated student (temp = 1.00) respectively. Document the following in the provided table:
            \begin{enumerate}
                \item Faithful Revisions (number of recommended revisions that the student correctly implements in the revised essay)
                \item Recommended Revisions Not Accurately Implemented and Unfaithful Revisions
                \begin{enumerate}
                    \item  \underline {Ignored Changes}: The student does not make any revisions to address the suggestion.
            
                    \item  \underline{Misinterpreted Changes}: The student attempts to address the suggestion, but the student’s actual revision differs or deviates from the feedback. In this case, the intent of the revision is aligned with the provided suggestion.
                    
                    \item  \underline{Inadequate Changes}: The student attempts to implement the suggestion but fails to address some explicit directive parts in the feedback such that the revision is highly insufficient to address the issue raised in the feedback or significantly impairs the original meaning of the suggestion.
                    
                    \item  \underline{Unfaithful Revisions}: The student introduces a substantial revision that does not correspond to any provided suggestions (i.e. recommended changes). Contrary to misinterpreted and inadequate changes, the intent of the change does not aligns with any of the suggestions.
                \end{enumerate}
            \end{enumerate}

            % \item Look through the suggestion list item by item to determine if the feedback is implemented correctly and fill out the table.

            % Example:
            % \begin{itemize}
            %     \item {\small Suggestion1: discuss consumer/producer surplus \checkmark \newline
            %     (Faithful: “This shift after the tax, and changes to producer surplus would then create a significant deadweight loss area, where both producer, and consumer lose out on combined surplus.”)} 
            %     \item {\small Suggestion2: discuss equilibrium \texttimes \newline
            %     (Ignored: No attempt made.)}
            %     \item {\small Suggestion3: discuss effect of min wage on supply/demand of automation market \texttimes \newline
            %     (Ignored: No attempt made.)}
            % \end{itemize}
            % Faithful Revisions: 1
            
            % Ignored Changes: 2

            % Misinterpreted Changes: 0
            
            % Inadequate Changes: 0
            
            % Unfaithful Revisions: 0
         \end{enumerate}\\
         \bottomrule
    \end{tabular}
\end{table*}

\end{document}